\documentclass[letterpaper,10pt,journal,twoside]{IEEEtran}

\usepackage[T1]{fontenc}
\usepackage[english]{babel}
\addto{\captionsenglish}{}
\usepackage{amsmath}
\usepackage{amsfonts}
\usepackage{amssymb}
\usepackage{amsthm}
\usepackage{bm}
\usepackage{xspace}
\usepackage[boxruled]{algorithm2e}
\usepackage{mathrsfs}
\usepackage{multirow}
\usepackage{eurosym}
\usepackage{dsfont}
\usepackage[normalem]{ulem} 
\usepackage{balance}
\usepackage{soul}
\usepackage{pdfpages}
\usepackage{graphicx}
\usepackage{todonotes}
\usepackage{url}
\usepackage{siunitx}
\usepackage{pgfplots}
\pgfplotsset{compat=1.18}
\usepackage{stfloats}
\usepackage{array}
\usepackage{adjustbox}
\usepackage{float}
\usepackage{enumitem}
\usepackage[normalem]{ulem}
\setitemize{leftmargin=*}
\allowdisplaybreaks
\usepackage{cite}
\usepackage{multibib}
\newcites{supp}{Supplementary References}
\usepackage{xfp}

\usepackage{environ}
\NewEnviron{myhideenv}{%
\setbox0\vbox{%
\let\xxwrite\write
\protected\def\write{\immediate\xxwrite}%
{\tiny XX\BODY XX}}
}

\usepackage{microtype}
\usepackage[subtle,tracking=normal]{savetrees}
\usepackage[hidelinks]{hyperref}

\newcolumntype{L}[1]{>{\raggedright\let\newline\\\arraybackslash\hspace{0pt}}m{#1}}
\newcolumntype{C}[1]{>{\centering\let\newline\\\arraybackslash\hspace{0pt}}m{#1}}
\newcolumntype{R}[1]{>{\raggedleft\let\newline\\\arraybackslash\hspace{0pt}}m{#1}}

\newcommand{\T}{\mathsf{T}}
\newcommand{\up}[1]{^{\text{#1}}}
\newcommand{\down}[1]{_{\text{#1}}}
\newcommand{\norm}[1]{\left\lVert#1\right\rVert}
\newcommand{\R}{\mathbb{R}}
\newcommand{\N}{\mathbb{N}}
\newcommand{\SE}{SE(3)}
\newcommand{\mat}[1]{\begin{bmatrix}#1\end{bmatrix}}
\newcommand{\fracdiff}[2]{\frac{\partial #1}{\partial #2}}

\definecolor{lightyellow}{rgb}{1.0,0.98,0.7}
\sethlcolor{lightyellow}
\newcommand{\trimscale}[1]{\fpeval{#1 * 0.3}}

\newcommand{\refsupp}[1]{#1 in supplementary material}

\title{
Multi-Contact Whole-Body Force Control\\for Position-Controlled Robots
}
\author{Quentin Rouxel, Serena Ivaldi, and Jean-Baptiste Mouret%
\thanks{Manuscript received: December, 29, 2023; Revised March, 16, 2024; Accepted April, 16, 2024. This paper was recommended for publication by Editor A. Kheddar upon evaluation of the Associate Editor and Reviewers' comments. This research was supported by the CPER CyberEntreprises, the Creativ’Lab platform of Inria/LORIA, the EurROBIN Horizon project (grant number 101070596), and the ANR in the France 2030 program through project PEPR O2R AS3 (ANR-22-EXOD-007).}%
\thanks{
Q. Rouxel, S. Ivaldi, and J.B. Mouret are with Inria, CNRS, Universit\'e de Lorraine, France. {\tt\footnotesize firstname.lastname@inria.fr}}%
\thanks{Digital Object Identifier (DOI): \url{https://doi.org/10.1109/LRA.2024.3396094}.}%
}

\begin{document}

% make the title area
\maketitle

%%%
%%% Abstract and keywords
%%%

\begin{abstract}
%Many humanoid and multi-legged robots are controlled in position, and not in torque, but this prevents them from having direct control over their contact forces. In turn, this prevents them from exploiting many contact with their environment, like putting a hand on a wall or a handrail, whereas it would critically improve their stability. In this paper, we take inspiration from the flexibility models used with serial elastic actuators to indirectly control contact forces on traditional position-controlled robots. We propose a two stages pipeline named SEIKO (Sequential Equilibrium Inverse Kinematic Optimization), performing whole body retargeting from Cartesian commands and admittance control through two quadratic programs (QP) formulation run at real time. We validated our pipeline with experiments on the real, full-scale humanoid robot Talos in various multi-contact scenarios, including pushing tasks, far-reaching tasks, stair climbing, and stepping on sloped surfaces. This work opens the possibility of stable, contact-rich behaviors while getting around many of the challenges of torque-controlled robots.

Many humanoid and multi-legged robots are controlled in positions rather than in torques, which prevents direct control of contact forces, and hampers their ability to create multiple contacts to enhance their balance, such as placing a hand on a wall or a handrail. This paper introduces the SEIKO (Sequential Equilibrium Inverse Kinematic Optimization) pipeline, and proposes a unified formulation that exploits an explicit model of flexibility to indirectly control contact forces on traditional position-controlled robots. SEIKO formulates whole-body retargeting from Cartesian commands and admittance control using two quadratic programs solved in real time. Our pipeline is validated with experiments on the real, full-scale humanoid robot Talos in various multi-contact scenarios, including pushing tasks, far-reaching tasks, stair climbing, and stepping on sloped surfaces. Code and videos are available at: \url{https://hucebot.github.io/seiko_controller_website/}

%In multi-contact, achieving configuration control for purely rigid position-controlled robots is challenging due to multiple possible contact force distributions existing for the same posture. This work includes an explicitly joint flexibility model into an optimization-based whole-body admittance controller for position-controlled robots. Building upon previous work SEIKO (Sequential Equilibrium Inverse Kinematic Optimization) for teleoperation retargeting, this formulation allows the realization of desired contact force distributions and enhances robustness against model inaccuracies. The system's robustness is evaluated through simulations and practical experiments with the Talos humanoid robot in various multi-contact scenarios, including pushing tasks, far-reaching tasks, stair climbing, and stepping on sloped surfaces.

\end{abstract}

%%%
%%% Document content
%%%

\section{Introduction}

\IEEEPARstart{H}{umans} often use additional contact points to enhance their stability, for instance, by using a handrail or a wall when walking, or to extend their reach, for instance, when grasping a distant object. While humanoid robots would benefit from a similar strategy, current robots minimize the number of contacts and use them only for feet and required interactions with the environment, such as pushing a button \cite{atkeson2018happened}.

The primary challenge in controlling multi-contact lies in the redundancy of force distribution resulting from closed kinematic chains \cite{featherstone2014rigid}. For a given posture with several contacts, there are infinite ways to distribute force among them. For instance, a humanoid with both hands on a table can apply more or less force to the hands without any visible change in joint position.

To regulate forces, most prior studies on multi-contact whole-body control rely on torque-controlled robots with inverse dynamics controllers \cite{cisneros2020inverse, henze2016passivity, abi2019torque}. Unfortunately, inverse dynamics is highly sensitive to model and calibration errors, and identifying models for humanoids is particularly challenging \cite{ramuzat2020actuator}. Perfect identification of environment's properties is generally not possible. This is why most deployed robots use position control, which is simpler and more reliable \cite{romualdi2020benchmarking}, but it lacks direct control authority over contact forces.
%, thus hindering the exploitation of multi-contact strategies.

Prior work on position-controlled robots \cite{kajita2010biped, caron2019stair, cisneros2019qp, samadi2021humanoid} has often regulated contact forces indirectly using various forms of admittance schemes applied independently to each effector. While effective in many scenarios, this strategy may lack robustness in challenging situations near physical limits or with significant model errors. This is due to its heuristic nature, which lacks theoretical grounding and fails to consider the whole-body effect of postural changes on contact forces.

\begin{figure}[t]
    \centering
    \includegraphics[trim=0cm 1.2cm 0cm 0cm,clip,width=\linewidth]{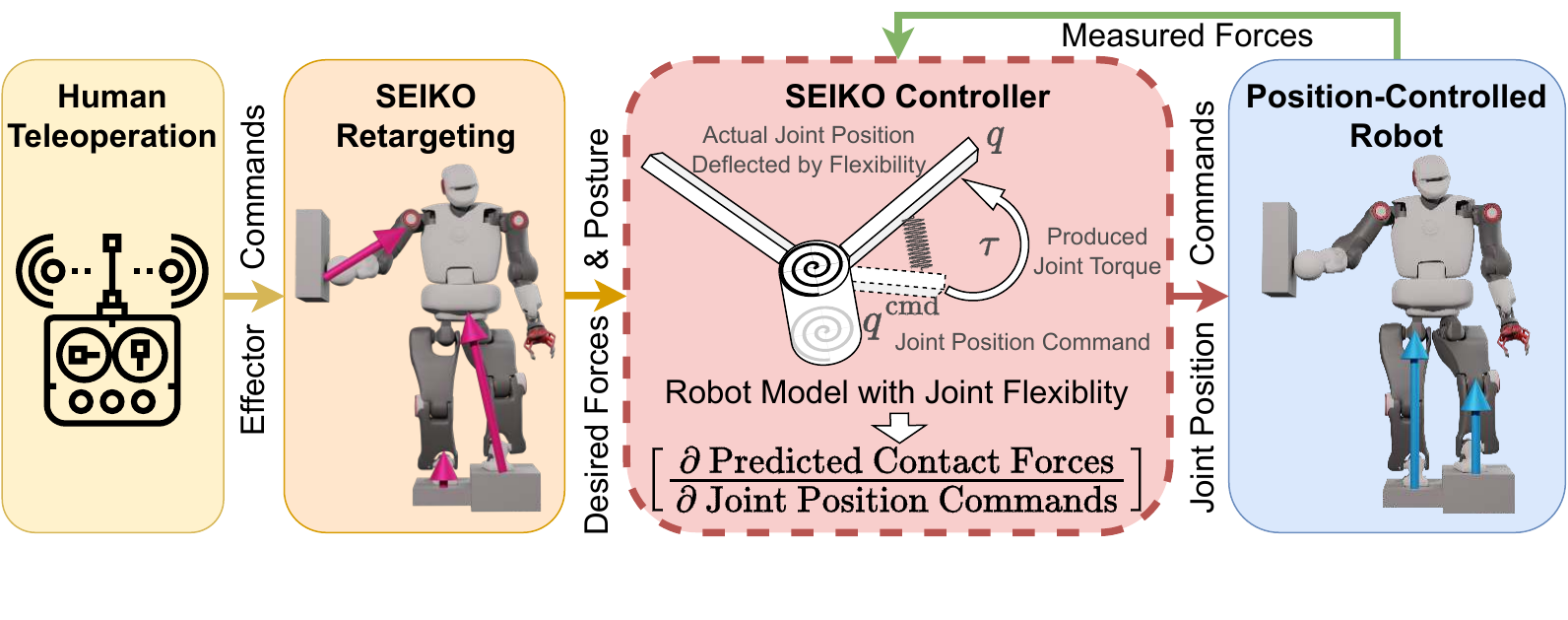}
    \includegraphics[trim=\trimscale{10}cm 0cm \trimscale{27}cm 0cm,clip,height=3.8cm]{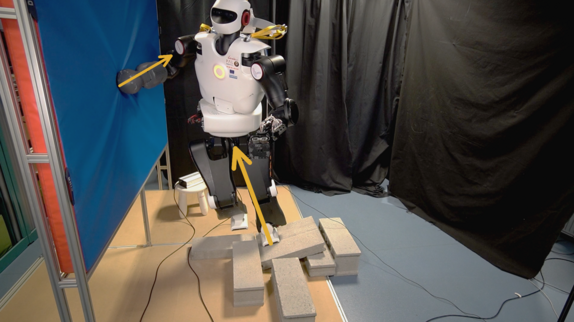}%
    \includegraphics[trim=\trimscale{7}cm 0cm \trimscale{32}cm 0cm,clip,height=3.8cm]{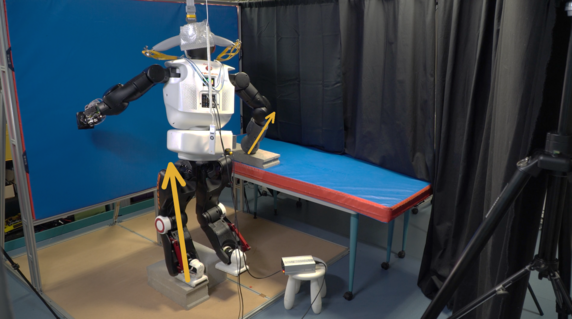}%
    \includegraphics[trim=\trimscale{18}cm 0cm \trimscale{21}cm 0cm,clip,height=3.8cm]{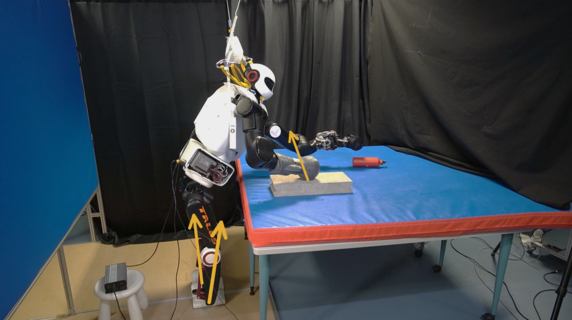}%
    \caption{Overview of our control pipeline (top), and illustrations of teleoperated multi-contact experiments on Talos humanoid robot (bottom).}
    \label{fig:concept}
\end{figure}

Our main idea is to exploit the robot's non-rigidity to explicitly model the relationship between joint position commands and contact forces. Flexibility arises from either non-observable mechanical structural bending or internal impedance of non-ideal joint position control. We present a control pipeline (Fig.~\ref{fig:concept}) designed to regulate contact forces on position-controlled robots. Our approach offers a novel unified whole-body formulation using optimization-based Quadratic Programming (QP) to leverage fast QP solvers.

We conducted experiments on the Talos humanoid robot \cite{stasse2017talos}, equipped with powerful arms but known for significant hip mechanical flexibility \cite{villa2022torque}. Our control pipeline is compatible with commands from autonomous planners and teleoperation, with a focus on the latter in this study. Well-suited for teleoperation, our method is robust against operator errors related to awareness and embodiment challenges. Unlike most existing methods, our approach enables motions close to feasibility boundaries (both in term of kinematic, balance, and torque limits), allowing full exploitation of the capabilities of the hardware.

Our work named SEIKO for Sequential Equilibrium Inverse Kinematic Optimization provides the following contributions:
\begin{itemize}    
    \item A Sequential QP (SQP) formulation that computes posture deflection and joint command correction, accounting for joint flexibility in multi-contact quasi-static conditions.
    \item A multi-contact retargeting and control architecture for position-controlled robots with contact switch and pushing capabilities, designed to be robust against model errors.
    \item Validation on the hardware Talos humanoid robot with several multi-contact tasks, including the validation of our prior retargeting work, which was previously tested only in simulation for humanoid robots.
\end{itemize}

\section{Related Work}\label{sec:related_works}

Multi-contact tasks have been studied in-depth on humanoid robots with torque control \cite{cisneros2020inverse,henze2016passivity, abi2019torque}, where contact forces are directly regulated using whole-body inverse dynamic controllers. But torque control relies on an accurate model of the robot's dynamic, which is challenging to identify \cite{ramuzat2020actuator, villa2022torque} and lacks robustness. Joint impedance control \cite{ruscelli2020multi, polverini2020multi} offers a more robust alternative to torque control, but still requires modeling the actuators to be able to specify torque feedforward references.

Although \cite{vaillant2016multi} demonstrated ladder climbing on a position-controlled robot without regulating contact forces, it is essential to regulate these forces. Doing so enables pushing tasks, smooth contact transitions, and enhances robustness for motions near system limits, where stability margins are reduced. Most studies using position-controlled robots \cite{caron2019stair, cisneros2019qp, samadi2021humanoid, murooka2022centroidal} regulate contact forces indirectly through methods named ``effector admittance'', ``foot force difference control'', or ``damping control'' all based on the same principle introduced in \cite{kajita2010biped}: an admittance feedback law is applied to each effector to adjust its Cartesian pose reference. For example, to reduce the force measured on the hand, these approaches will retract its desired position away from the contact surface. However, because it is expected that the hand remains in contact, these approaches implicitly rely on flexibility without explicitly considering it. In contrast, SEIKO Controller models the whole-body flexibility, providing a more grounded formulation. This also enables the accounting of postural changes on the contact forces, which the effector admittance scheme cannot do.

\cite{khatib2008torque} proposed the idea of modeling torques produced by position-controlled actuators, which was further studied in \cite{del2016implementing} and applied to multi-contact in \cite{farnioli2016toward, hiraoka2021online}. Similar to our approach, they differentiate the quasi-static equilibrium but their method uses pseudo-inverses which fails at considering constraints. Furthermore, \cite{hiraoka2021online} also uses elastic joint models, but their method solves a cascade of several QP problems. Their purely reactive control architecture lacks feedforward terms and retargeted references, making it more sensitive to noise and violations of the quasi-static assumption. In contrast, our method is unified, allows faster motions, and does not require actual joint positions to be measured, which accommodates robots with mechanical flexibility like the Talos robot.

This work builds upon our prior work \cite{seiko, seiko2, seiko3} which devised a retargeting framework for multi-contact tasks on simulated humanoids and hardware bimanual manipulators with an highlight on enforcing feasibility. Both this work and \cite{seiko, seiko2, seiko3} also target teleoperation applications. While many studies have investigated the teleoperation of complex robots with floating bases \cite{darvish2023teleoperation}, fewer have explicitly addressed multi-contact scenarios \cite{di2016multi, mccrory2023generating}. In contrast to these, our work addresses the regulation of contact forces and demonstrates both contact switch and pushing tasks.

\section{Problem Definition}\label{sec:challenges}

\begin{table}[t]
\centering
\caption{Mathematical notations}
\label{table:notations}
\begin{tabular}{|cl|} 
    \hline
    Notation & Description\\
    \hline
    $n \in \N$ & Number of joints\\
    $m\down{plane} \in \N$ & Number of enabled plane contacts\\
    $m\down{point} \in \N$ & Number of enabled point contacts\\
    $m = 6m\down{plane} + 3m\down{point}$ & Dimension of stacked wrench\\
    $(\bullet)\up{read}$ & Estimated measured quantities\\
    $(\bullet)\up{op}$ & Operator's raw Cartesian commands\\
    $(\bullet)\up{adm}$ & Effectors admittance scheme quantities\\
    $(\bullet)\up{target}$ & Processed commands for retargeting input\\
    $(\bullet)\up{d}$ & Desired state computed by retargeting\\
    $(\bullet)\up{flex}$ & Flexible state computed by controller\\
    $(\bullet)\down{on}$ & Enabled contact quantities\\
    $(\bullet)\down{off}$ & Disabled contact (free effector) quantities\\
    $\bm{X} \in \SE$ & Cartesian pose\\ 
    $\bm{\nu} \in \R^6$ & Cartesian spatial velocity\\
    $\bm{q} \in \R^{7+n}$ & Posture position (floating base and joints)\\
    $\dot{\bm{q}} \in \R^{6+n}$ & Posture velocity\\
    $\bm{\theta},\dot{\bm{\theta}} \in \R^n$ & Joint position and velocity\\
    $\bm{\theta}\up{cmd} \in \R^n$ & Joint position command sent to robot\\
    $\bm{\theta}\up{min}, \bm{\theta}\up{max} \in \R^n$ & Joint position min/max bounds\\
    $\Delta\bm{\lambda}\up{effort} \in \R^m$ & Wrench effort (input to controller)\\
    $\bm{\tau} \in \R^n$ & Joint torque\\
    $\bm{\tau}\up{max} \in \R^n$ & Absolute maximum joint torque\\
    $\tilde{\bm{\tau}}\up{max} \in \R^n$ & Joint torque limits used in Controller\\
    $\bm{\lambda} \in \R^m$ & Stacked contact wrench\\
    $\bm{k} \in \R^n$ & Joint stiffness vector\\
    $\bm{K} = \mathsf{diag}(\bm{k}) \in \R^{n \times n}$ & Joint stiffness matrix\\
    $\bm{S} \in \R^{(6+n) \times n}$ & Selection matrix joint to full dimension\\
    $\bm{S'} \in \R^{n \times (6+n)}$ & Selection matrix full to joint dimension\\
    $\bm{g}(\bm{q}) \in \R^{6+n}$ & Gravity vector\\
    $\bm{J}\down{on/off}(\bm{q}) \in \R^{(6+n) \times m}$ & Stacked effectors Jacobian matrix\\
    $K_p, K_d \in \R$ & Proportional and derivative control gains\\
    $K\down{adm}\in \R$ & Effectors admittance gain\\
    $\Delta t \in \R$ & Time step\\
    $\mathsf{FK}\down{on/off}(\bm{q})$ & Effector poses (forward kinematic)\\
    $\oplus, \ominus$ & Operations on $\SE$ Lie algebra\\
    \hline
\end{tabular}
\end{table}

Quasi-static robot configurations are defined by postural positions, joint torques, and contact wrenches $\bm{q}, \bm{\tau}, \bm{\lambda}$. For position-controlled robots, control inputs only consist of joint position commands $\bm{\theta}\up{cmd}$. The whole-body retargeting stage (illustrated in Fig.~\ref{fig:concept} and proposed in previous work \cite{seiko}) provides a stream of desired quasi-static configurations $\bm{q}\up{d},\bm{\tau}\up{d},\bm{\lambda}\up{d}$ expected to be feasible.

Achieving desired contact wrenches $\bm{\lambda}\up{d}$ is essential for multi-contact tasks, but contact wrenches can not be directly commanded on position-controlled robots. Our approach aims to indirectly control contact wrenches through joint position commands $\bm{\theta}\up{cmd}$ optimized to take into account the flexibility of the robot. Table~\ref{table:notations} lists the notations and quantities used throughout this letter.

Addressing the problem involves overcoming the following challenges:
\begin{itemize}
    \item Multi-contact tasks exhibit redundancy in both kinematics and contact wrench distribution, akin to the Grasp matrix's nullspace in manipulation \cite{rimon2019mechanics}.
    \item While adding contacts is generally feasible, removing contacts challenge the robot's balance and can be infeasible.
    \item Transitioning between contact states (enabled or disabled) involves discrete changes in problem formulation. Ensuring continuity in contact wrenches (from non-zero to zero and vice versa) and posture is essential for smooth transitions.
    \item To ensure safety, physical limits must be enforced such as balance, joint kinematics, actuator torque limits, and contact stability conditions prohibiting pulling, sliding, tilting.
    \item For application to hardware, the controller must be robust to model errors and violations of simplifying assumptions.
\end{itemize}

\section{Method}

\begin{figure*}[t]
    \centering
	\includegraphics[trim=0cm 0cm 0cm 0cm,clip,width=\linewidth]{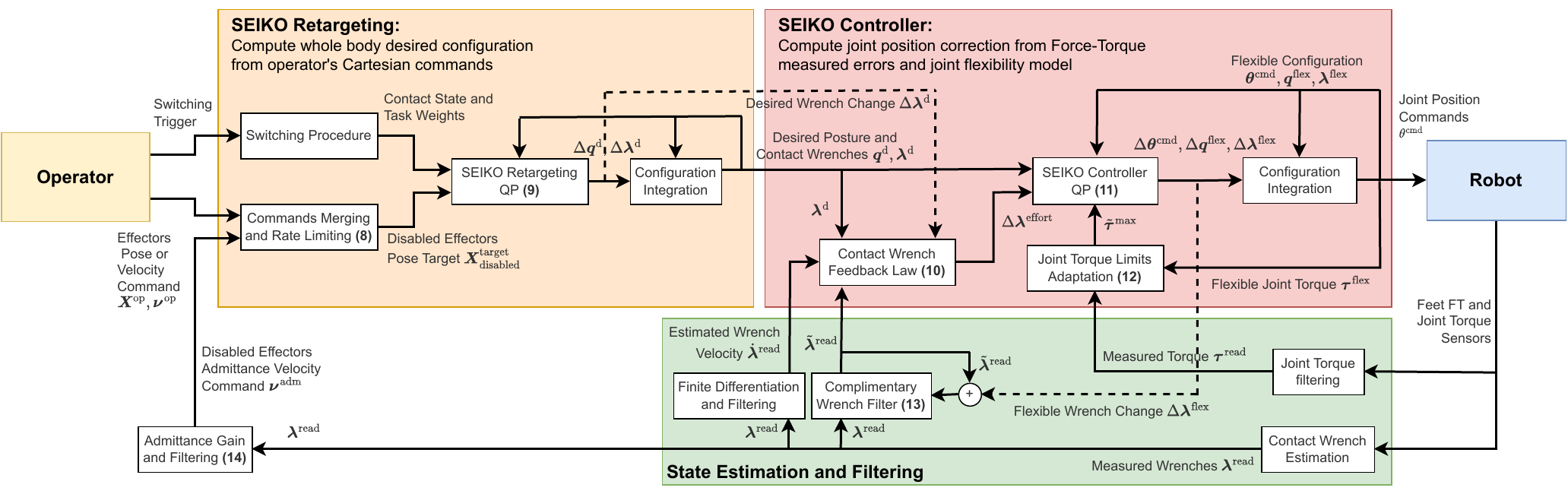}	
    \caption{Control architecture for position-controlled robots: Operator's Cartesian commands are retargeted into a feasible whole-body configuration. The controller uses a joint flexibility model to adjust actuator position commands for contact wrench control and prevent exceeding joint torque limits.}
    \label{fig:architecture}
\end{figure*}

\subsection{Main Idea}

According to rigid body theory in multi-contact \cite{featherstone2014rigid, rimon2019mechanics}, the contact wrenches of an ideal infinitely stiff mechanical system are non-unique and lie in a redundant nullspace. Real systems, however, always exhibit inherent flexibility: the structure slightly bends, and both the deflected posture and contact wrenches uniquely evolve towards the configuration minimizing overall elastic energy. Therefore, given constant joint position commands, the mapping that takes into account flexibility $\bm{\theta}\up{cmd} \mapsto (\bm{q}\up{flex}, \bm{\lambda}\up{flex})$ is unique and well-defined. Our approach models and predicts this whole-body non-linear deflection effect, utilizing it for the control of contact wrenches.

Specifically, we linearize and compute derivatives of the deflection effect to consider how contact wrenches change with variations in joint position commands through the Jacobian matrix $\frac{\partial \bm{\lambda}\up{flex}}{\partial \bm{\theta}\up{cmd}}(\bm{q}\up{flex}, \bm{\lambda}\up{flex}, \bm{\theta}\up{cmd})$. Instead of directly inverting this Jacobian matrix, we formulate the control problem as a Quadratic Programming (QP) which solves for position command changes and optimizes multiple objectives, similar to task space inverse dynamic approaches. We explicitly model the system's flexibility by treating each robot joint as a spring, encompassing both internal actuator impedance and mechanical flexibilities.

\subsection{Overall Architecture}

Our proposed control architecture depicted in Fig.~\ref{fig:architecture}  consists of a two-stage pipeline. Firstly, \textit{SEIKO Retargeting}, previously introduced in \cite{seiko}, optimizes a desired whole-body configuration $\bm{q}\up{d}, \bm{\lambda}\up{d}, \bm{\tau}\up{d}$ within feasibility limits. Subsequently, our novel \textit{SEIKO Controller} computes corrected joint position commands $\bm{\theta}\up{cmd}$ for tracking $\bm{\lambda}\up{d}$. These joint commands are then sent to the robot's low-level servomotors and tracked by stiff internal position controllers.

The controller has three goals: (i) achieve the desired contact wrenches $\bm{\lambda}\up{d}$, (ii) avoid violations of joint torque limits $\bm{\tau}\up{max}$, and (iii) enhance robustness against model inaccuracies. The Retargeting step is crucial as it enforces feasibility limits a priori, and generates a desired configuration to be tracked. The controller indeed exhibits reduced stability when tracking a highly infeasible non-retargeted reference.

The set of effectors that may come into contact with the environment is pre-defined. Each effector's state is either: ``enabled'', standing for fixed and in contact transmitting forces and torques to the environment, or ``disabled'', indicating that it is free to move and is commanded by the operator. Our formulation handles both plane contacts (6 DoFs, e.g., feet) and point contacts (3 DoFs, e.g., hands with ball shape). The full details of contact formulation are available in the supplementary material of \cite{seiko}. Other types of contacts can also be easily implemented, such as full grasp contact for hand grippers or even line contact on the edge of feet.

An external planner or human operator provides commands as input to the Retargeting stage: (i) Cartesian pose $\bm{X}\up{op}$ or velocity $\bm{\nu}\up{op}$ commands for each free (disabled) effector, (ii) a Boolean signal that manually triggers the transition between contact states, and (iii) an optional ``pushing mode'' enabling explicit control of the normal force of a specific enabled contact. Our method does not plan contact sequencing, and relies on external decisions for contact stances and sequence.

The proposed method operates instantaneously without considering the future of unknown intention, and relies on the quasi-static assumption. The nonlinear whole-body optimizations are solved using SQP schemes with only one QP iteration per time step. This allows for quick convergence at high frequency (\SI{500}{\hertz}) and responsiveness to input changes.

\subsection{Equilibrium Equation and Flexibility Model}

Motions of mobile robots with a floating base are governed by the equation of motion in joint space \cite{featherstone2014rigid}. Under the quasi-static assumption, where $\ddot{\bm{q}} \approx \dot{\bm{q}} \approx \bm{0}$, this equation simplifies to represent the equilibrium, i.e. system's balance, between contact wrenches, gravity effects, and applied torques:
\begin{equation}\label{eq:equilibrium}
    \bm{g}(\bm{q}) = \bm{S}\bm{\tau} + \bm{J}(\bm{q})^\T \bm{\lambda},
\end{equation}
which is non-linear in $\bm{q}$. We approximate the linearization of the equilibrium equation by considering small variations of the configuration $\left(\bm{q}+\Delta\bm{q}, \bm{\lambda}+\Delta\bm{\lambda}, \bm{\tau}+\Delta\bm{\tau}\right)$ and partial derivatives:
\begin{multline}\label{eq:diff_equilibrium}
    \bm{g}(\bm{q}) + \fracdiff{\bm{g}}{\bm{q}}\Delta\bm{q} =
    \bm{S}\bm{\tau} + \bm{S}\Delta\bm{\tau} \\
    + \bm{J}(\bm{q})^\T\bm{\lambda} + \bm{J}(\bm{q})^\T\Delta\bm{\lambda} 
    + \left(\fracdiff{\bm{J}}{\bm{q}}^\T\bm{\lambda}\right)\Delta\bm{q},  
\end{multline}
while neglecting second order terms (see \refsupp{Section~{\ref{sec:supp_equations}}} for details).

Stiff position-controlled robots deviate from the rigid assumption due to inherent hardware flexibility arising from factors like Series Elastic Actuators \cite{nava2017momentum}, deformations in links or transmissions \cite{villa2022torque}, impedance of non-ideal position control \cite{khatib2008torque}, or the inclusion of soft damper elements within the structure \cite{koch2014optimization}. In this work, we model this flexibility as joint elastic flexibility, where the relation between joint position and generated torque is expressed as follows:
\begin{equation}
    \bm{\tau}\up{flex} = \bm{K}(\bm{\theta}\up{cmd} - \bm{\theta}\up{flex}).
\end{equation}
Note that link flexibility can also be modeled in a similar manner by introducing passive joints without actuation. Its derivative is written:
\begin{equation}\label{eq:diff_flexiblity}
    \Delta\bm{\tau}\up{flex} = \bm{K}(\Delta\bm{\theta}\up{cmd} - \Delta\bm{\theta}\up{flex}) = \bm{K}\left(\Delta\bm{\theta}\up{cmd} - \bm{S'}\Delta\bm{q}\up{flex}\right),
\end{equation}
where $\bm{q}\up{flex}$ is the deflected posture under joint flexibility and $\bm{\theta}\up{cmd}$ is the joint position command of actuators.

The derivative-based linear approximation of the equilibrium equation (\ref{eq:diff_equilibrium}) combined with flexibility model (\ref{eq:diff_flexiblity}) is linear w.r.t. configuration changes:
%\begin{equation}
%\begin{aligned}
%& \bm{g}(\bm{q}\up{flex}) + \fracdiff{\bm{g}}{\bm{q}}(\bm{q}\up{flex})\Delta\bm{q}\up{flex} = 
%~\bm{S}\bm{\tau}\up{flex}\\
%&+ \bm{S}\bm{K}\Delta\bm{\theta}^{\text{cmd}} - \bm{S}\bm{K}\bm{S'}\Delta\bm{q}\up{flex}\\
%&+ \bm{J}(\bm{q}\up{flex})^\T\bm{\lambda}\up{flex} + \bm{J}(\bm{q}\up{flex})^\T\Delta\bm{\lambda}\up{flex}\\
%&+ \left( \fracdiff{\bm{J}}{\bm{q}}^\T(\bm{q}\up{flex})\bm{\lambda}\up{flex} \right) \Delta\bm{q}\up{flex}
%\end{aligned}
%\end{equation}
\begin{equation}\label{eq:diff_equilibrium_flexiblity}
\begin{aligned}
    & \bm{S}\bm{K}\Delta\bm{\theta}\up{cmd} = \bm{T}(\bm{q}\up{flex},\bm{\lambda}\up{flex})\mat{\Delta\bm{q}\up{flex} \\ \Delta\bm{\lambda}\up{flex}} + \bm{t}(\bm{q}\up{flex},\bm{\lambda}\up{flex},\bm{\theta}^{\text{cmd}}) \\
    & \text{where } \bm{T}(\bm{q}\up{flex},\bm{\lambda}\up{flex}) = \\
    & \mat{\fracdiff{\bm{g}}{\bm{q}}(\bm{q}\up{flex})-\left(\fracdiff{\bm{J}}{\bm{q}}^\T(\bm{q}\up{flex})\bm{\lambda}\up{flex}\right)+\bm{S}\bm{K}\bm{S'} & |~-\bm{J}(\bm{q}\up{flex})^\T},\\
    & \bm{t}(\bm{q}\up{flex},\bm{\lambda}\up{flex},\bm{\theta}^{\text{cmd}}) = \bm{g}(\bm{q}\up{flex}) - \bm{S}\bm{\tau}\up{flex} - \bm{J}(\bm{q}\up{flex})^\T\bm{\lambda}\up{flex}.\\
\end{aligned}
\end{equation}
Therefore $\Delta\bm{\theta}\up{cmd}$ can also be linearly expressed from $\Delta\bm{q}\up{flex}$ and $\Delta\bm{\lambda}\up{flex}$ using the following row decomposition:
\begin{equation}\label{eq:diff_equilibrium_flexiblity_split}
    \mat{\bm{0} \\ \bm{K}\Delta\bm{\theta}\up{cmd}} = 
    \mat{
        \bm{T}\down{B} \\ \bm{T}\down{J}}
    \mat{\Delta\bm{q}\up{flex} \\ \Delta\bm{\lambda}\up{flex}} + 
    \mat{\bm{t}\down{B} \\ \bm{t}\down{J}}
\end{equation}
\begin{equation}\label{eq:flex_cmd_mapping}
    \Delta\bm{\theta}\up{cmd}(\Delta\bm{q}\up{flex}, \Delta\bm{\lambda}\up{flex}) = \bm{K}^{-1}\left(\bm{T}\down{J}\mat{\Delta\bm{q}\up{flex} \\ \Delta\bm{\lambda}\up{flex}} + \bm{t}\down{J}\right),
\end{equation}
where $\bm{T}\down{B}, \bm{t}\down{B}$ refer to the first 6 rows representing the floating base and $\bm{T}\down{J}, \bm{t}\down{J}$ the remaining $n$ joint rows.

\subsection{SEIKO Retargeting}

This section summarizes the SEIKO Retargeting method developed in \cite{seiko, seiko2}. From this previous work, \refsupp{Section~{\ref{sec:supp_balance_conditions}}} provides further explanation on how balance is enforced.

The Retargeting preprocesses inputs for each disabled effector, which includes the commanded motion from the operator (comprising both pose $\bm{X}\up{op}$ and velocity $\bm{\nu}\up{op}$) and the admittance velocity command $\bm{\nu}\up{adm}$ (see Section \ref{sec:admittance}). Processing includes filtering and merging these commands:
\begin{equation}\label{eq:cmd_filtering}
\begin{aligned}
    & \bm{X}\up{target}\down{off} = \mathsf{filtering} \left( \bm{X}\down{ref}(t)\bm{X}\up{op} \right), ~\bm{X}\down{ref}(0) = \bm{X}\up{read}, \\
    & \bm{X}\down{ref}(t+\Delta t) = \\
    & \mathsf{boundDistance}\left(\bm{X}\down{ref}(t) \oplus \Delta t(\bm{\nu}\up{op} + \bm{\nu}\up{adm}), ~\bm{X}\up{d}\down{off} \right),
\end{aligned}
\end{equation}
where $\bm{X}\down{ref} \in \SE$ is a reference pose that integrates velocity commands at each time step (see \cite{seiko2}). $\bm{X}\down{ref}$ allows the Cartesian pose command $\bm{X}\up{op}$ to be expressed relative to this reference, and not in arbitrary world frame. The filtering process incorporates a smoothing low-pass filter and bounds signal's velocity and acceleration through time-optimal bang-bang trajectory replanning \cite{seiko2}. With the clamping $\mathsf{boundDistance}()$, we also constrain separately the position and orientation of $\bm{X}\down{ref}$ within a radius centered on $\bm{X}\up{d}\down{off}$ to prevent the reference pose from windup when the retargeted motion is saturated by the feasibility constraints.

At each time step, SEIKO Retargeting solve the QP:
\begin{subequations}
\begin{align}
    \underset{\Delta\bm{\bm{q}}\up{d}, \Delta\bm{\bm{\lambda}}\up{d}, \Delta\bm{\bm{\tau}}\up{d}}{\text{argmin}} & \label{eq:retargeting_min}\\
    & \norm{\mathsf{FK}\down{off}(\bm{q}\up{d}) \oplus \bm{J}\down{off}(\bm{q}\up{d})\Delta\bm{q}\up{d} \ominus \bm{X}\down{off}\up{target}}^2 + \label{eq:retargeting_cost_cart}\\
    & \norm{\bm{\theta}\up{d} + \Delta\bm{\theta}\up{d} - \bm{\theta}\up{target}}^2 + \label{eq:retargeting_cost_joint}\\
    & \norm{\bm{\tau}\up{d} + \Delta\bm{\tau}\up{d}}^2 + \label{eq:retargeting_cost_tau}\\
    & \norm{\bm{\lambda}\up{d} + \Delta\bm{\lambda}\up{d}}^2 + \label{eq:retargeting_cost_lambda}\\
    & \norm{\Delta\bm{q}\up{d}}^2 + \norm{\Delta\bm{\lambda}\up{d}}^2 \label{eq:retargeting_cost_reg}\\
    \text{such that}\notag\\
    & \text{linearized equilibrium equation (\ref{eq:diff_equilibrium})} \label{eq:retargeting_eq_equilibrium}\\
    & \mathsf{FK}\down{on}(\bm{q}\up{d}) \oplus \bm{J}\down{on}(\bm{q}\up{d})\Delta\bm{q}\up{d} \ominus \bm{X}\up{target}\down{on} = \bm{0} \label{eq:retargeting_eq_cart}\\
    & \bm{\theta}\up{min} \leq \bm{\theta}\up{d} + \Delta\bm{\theta}\up{d} \leq \bm{\theta}\up{max} \label{eq:retargeting_ineq_joint}\\
    & -\bm{\tau}\up{max} \leq \bm{\tau}\up{d} + \Delta\bm{\tau}\up{d} \leq \bm{\tau}\up{max} \label{eq:retargeting_ineq_tau}\\
    & \bm{C}\down{contact}\Delta\bm{\lambda}\up{d} + \bm{C}\down{contact}\bm{\lambda}\up{d} + \bm{c}\down{contact} \geq \bm{0} \label{eq:retargeting_ineq_contact}\\
    & -\Delta t\dot{\bm{\theta}}\up{max} \leq \Delta\bm{\theta}\up{d} \leq \Delta t\dot{\bm{\theta}}\up{max} \label{eq:retargeting_ineq_vel_joint}\\
    & -\Delta t\dot{\bm{\lambda}}\up{max} \leq \Delta\bm{\lambda}\up{d} \leq \Delta t\dot{\bm{\lambda}}\up{max} \label{eq:retargeting_ineq_vel_lambda}.
\end{align}
\end{subequations}
The QP solves for the configuration change (\ref{eq:retargeting_min}), integrating it to update the desired configuration, e.g., $\bm{\lambda}\up{d}(t+\Delta t) = \bm{\lambda}\up{d}(t) + \Delta \bm{\lambda}\up{d}$. The optimization minimizes tasks weighted by manually tuned parameters for stability and desired trade-off. The cost function includes disabled effector pose targets (\ref{eq:retargeting_cost_cart}), default joint position targets (\ref{eq:retargeting_cost_joint}) for regularization and mitigating kinematic local minima, joint torque minimization (\ref{eq:retargeting_cost_tau}) for human-like postures, contact wrench penalization (\ref{eq:retargeting_cost_lambda}), and decision variable regularization (\ref{eq:retargeting_cost_reg}).

Equality constraints enforce the linearized equilibrium equation (\ref{eq:retargeting_eq_equilibrium}) and ensure enabled contacts are fixed (\ref{eq:retargeting_eq_cart}). Inequality constraints include joint position limits (\ref{eq:retargeting_ineq_joint}), joint torque limits (\ref{eq:retargeting_ineq_tau}), and contact stability conditions (\ref{eq:retargeting_ineq_contact}) considering unilaterality, friction pyramid, and center of pressure (see \refsupp{Section~{\ref{sec:supp_contact_conditions}}}). Additional constraints involve limits on joint changes (\ref{eq:retargeting_ineq_vel_joint}) and contact wrench changes (\ref{eq:retargeting_ineq_vel_lambda}).

%At initialization prior to the first iteration, we use a conventional Inverse Dynamic QP to reset the desired contact wrench and joint torque $\bm{\lambda}\up{d}, \bm{\tau}\up{d}$ from measured posture. This step prevents QP failures caused by initial large change and velocity constraints.

Compared to our prior SEIKO Retargeting work \cite{seiko}, we enhanced the contact switching procedure with fewer arbitrary choices and clearer physical semantics. Details can be found in \refsupp{Section~{\ref{sec:supp_contact_switch}}}.
%We enhanced the contact switching procedure compared to prior work. To remove a contact, we instantly increase the weight of the wrench penalty task to a very high value and use joint velocity $\dot{\bm{\theta}}\up{max}$ and wrench velocity $\dot{\bm{\lambda}}\up{max}$ limits to ensure a smooth transition. When the integrated desired wrench falls below a small threshold, the contact is removed. Enabling a contact is straightforward, as it doesn't require any special considerations, thanks to these limits.

\subsection{SEIKO Controller}

We assume that actual joint positions under flexibility cannot be directly measured but can be estimated from the model. Despite model errors, our approach relies on the model's derivatives direction to provide sufficient information about system evolution. The controller uses the derivative-based linear approximation of the equilibrium equation with flexibility (\ref{eq:diff_equilibrium_flexiblity}) to model how contact wrench distribution changes with joint command changes $\Delta \bm{\theta}\up{cmd}$. This approach generalizes previously used admittance control laws such as ``foot difference control'' \cite{kajita2010biped} which implicitly depends on flexibility without considering it.

The following feedback law regulates contact wrenches. It is the only feedback effect in our unified formulation that uses measured quantities and is tuned with only two parameters:
\begin{equation}\label{eq:feedback_law}
    \Delta\bm{\lambda}\up{effort} = \Delta\bm{\lambda}\up{d} + K_p(\bm{\lambda}\up{d} - \tilde{\bm{\lambda}}\up{read}) - K_d\dot{\bm{\lambda}}\up{read},
\end{equation}
where $\Delta\bm{\lambda}\up{effort}$ is the desired effort in the controller optimization, and $\Delta\bm{\lambda}\up{d}$ acts as a feedforward term. SEIKO Controller solves the following QP at each time step::
\begin{subequations}
\begin{align}
    \underset{\Delta\bm{\bm{q}}\up{flex}, \Delta\bm{\bm{\lambda}}\up{flex}}{\text{argmin}} & \label{eq:controller_min}\\
    & \norm{\Delta\bm{\lambda}\up{effort} - \Delta\bm{\lambda}\up{flex}}^2 + \label{eq:controller_cost_lambda}\\
    & \norm{\mathsf{FK}\down{off}(\bm{q}\up{flex}) \oplus \bm{J}\down{off}(\bm{q}\up{flex})\Delta\bm{q}\up{flex} \ominus \bm{X}\down{off}\up{d}}^2 +\label{eq:controller_cost_cart}\\
    & \norm{\bm{\theta}\up{cmd}+\Delta\bm{\theta}\up{cmd}-\bm{\theta}\up{d}}^2 + \label{eq:controller_cost_cmd}\\
    & \norm{\Delta\bm{\theta}\up{cmd}}^2\label{eq:controller_cost_cmd_delta}\\
    \text{such that}\notag\\
    & \bm{T}\down{B}\mat{\Delta\bm{q}\up{flex} \\ \Delta\bm{\lambda}\up{flex}}+\bm{t}\down{B} = \bm{0} \label{eq:controller_eq_equilibrium}\\
    & \mathsf{FK}\down{on}(\bm{q}\up{flex}) \oplus \bm{J}\down{on}(\bm{q}\up{flex})\Delta\bm{q}\up{flex}  \ominus \bm{X}\up{target}\down{on} = \bm{0} \label{eq:controller_eq_cart}\\
    & \bm{\theta}\up{min} \leq \bm{\theta}\up{cmd} + \Delta\bm{\theta}\up{cmd} \leq \bm{\theta}\up{max} \label{eq:controller_ineq_joint}\\
    & -\tilde{\bm{\tau}}\up{max} \leq \bm{\tau}\up{flex} + \Delta\bm{\tau}\up{flex} \leq \tilde{\bm{\tau}}\up{max} \label{eq:controller_ineq_tau}.
\end{align}
\end{subequations}
The QP solves for flexible configuration changes $\Delta\bm{\bm{q}}\up{flex}, \Delta\bm{\bm{\lambda}}\up{flex}$ (\ref{eq:controller_min}). Joint command changes $\Delta\bm{\theta}\up{cmd}$ are obtained from the decision variables using (\ref{eq:flex_cmd_mapping}) and $\bm{\bm{q}}\up{flex}, \bm{\bm{\lambda}}\up{flex}, \bm{\theta}\up{cmd}$ are then obtained by integration.

The cost function primarily computes joint position correction $\Delta\bm{\theta}\up{cmd}$ and resulting posture deflection $\Delta\bm{q}\up{flex}$ to achieve the control effort on contact wrench changes $\Delta\bm{\lambda}\up{effort}$ (\ref{eq:controller_cost_lambda}). It also adjusts disabled effector poses influenced by flexibility toward Retargeting's desired poses (\ref{eq:controller_cost_cart}). As secondary objectives, the optimization penalizes the discrepancy between corrected and desired joint positions (\ref{eq:controller_cost_cmd}) and regularizes changes in joint commands (\ref{eq:controller_cost_cmd_delta}).

Equality constraints enforce the linearized equilibrium equation with flexibility (\ref{eq:controller_eq_equilibrium}) through the first upper 6 floating base rows of decomposition (\ref{eq:diff_equilibrium_flexiblity_split}) and ensure no Cartesian motion for enabled contacts (\ref{eq:controller_eq_cart}). Inequality constraints ensure kinematic limits of joint position commands $\bm{\theta}\up{cmd}$ (\ref{eq:controller_ineq_joint}) and restrict maximum joint torques (\ref{eq:controller_ineq_tau}).

Joint torque limits $\tilde{\bm{\tau}}\up{max}$ used as constraints are dynamically updated to prevent the integrated state $|\bm{\tau}\up{flex}|$ from continuously increasing when the measured joint torque $|\bm{\tau}\up{read}|$ reaches the defined torque limit $\bm{\tau}\up{max}$. For each joint at each time step:
\begin{equation}\label{eq:limit_joint_tau_adaptation}
    \tilde{\tau}\up{max}(t+\Delta t) = 
    \begin{cases}
        \tau\up{flex}+\epsilon_1 & \text{if } 
        |\tau\up{read}|>\tau\up{max} ~\wedge \\
        & \tilde{\tau}\up{max}(t)>\tau\up{flex}+\epsilon_1, \\
        \tilde{\tau}\up{max}(t)+\epsilon_2 & \text{else if } 
        |\tau\up{read}|<\tau\up{max}-\epsilon_3 ~\wedge \\
        & \tilde{\tau}\up{max}(t) < \tau\up{max}, \\
        \tau\up{max} & \text{else if } \tilde{\tau}\up{max}(t) > \tau\up{max},\\
        \tilde{\tau}\up{max}(t) & \text{else},
    \end{cases}
\end{equation}
where $\epsilon_1, \epsilon_2, \epsilon_3 \in \R$ are small positive margin parameters implementing a hysteresis effect to improve stability. 

%At initialization and when a contact is removed, the flexible configuration $\bm{q}\up{flex},\bm{\lambda}\up{flex}$ is reset while keeping $\bm{\theta}\up{cmd}$ constant. This reset is achieved by running a few iterations of a flexible version of the Inverse Dynamic problem, also implemented as a SQP. This ensures an appropriate starting configuration and maintains smooth transitions during initialization and contact state changes. Such continuity is important as discrete changes in contact state can otherwise lead to discontinuities due to the flexibility model, especially when $\bm{\lambda}\up{flex}$ is not brought to zero due to modeling errors.

\subsection{State Estimation and Effectors Admittance}\label{sec:admittance}

%On our Talos humanoid robot, we use ankle-mounted Force-Torque sensors for foot contact wrench measurement ($\bm{\lambda}\up{read}\down{feet}$). To handle heavy loads in multi-contact scenarios, we have replaced the right forearm with a 3D-printed ball-shaped hand, rendering wrist Force-Torque sensors unusable. Hand contact forces are then indirectly estimated using arm joint torque measurements and robot kinematics: $\bm{f}\up{read}\down{hands} = \left(\bm{J}^\T(\bm{q}\up{read})\right)^{-1}(\bm{g}(\bm{q}\up{read}) - \bm{\tau}\up{read})$.\\

The estimated measured wrench $\tilde{\bm{\lambda}}\up{read}$ in feedback law (\ref{eq:feedback_law}) is computed using a complementary filter:
\begin{equation}
    \tilde{\bm{\lambda}}\up{read}(t+\Delta t) = \alpha \left( \tilde{\bm{\lambda}}\up{read}(t) + \Delta \bm{\lambda}\up{flex} \right) + (1-\alpha)\bm{\lambda}\up{read}.
\end{equation}
This filter enhances closed-loop stability by mitigating dynamical effects affecting $\bm{\lambda}\up{read}$ neglected by the quasi-static assumption. It introduces a trade-off between the reactive measurement and the term estimated through the integration of the predicted change $\Delta \bm{\lambda}\up{flex}$. The measured contact wrench velocity $\dot{\bm{\lambda}}\up{read}$ is computed using finite differences from $\bm{\lambda}\up{read}$, and then it is low-pass filtered at $10$ Hz using an exponential first-order scheme.

We utilize an admittance scheme to compute an additional Cartesian velocity command for disabled effectors $\bm{\nu}\up{adm}$:
\begin{equation}\label{eq:admittance}
    \bm{\nu}\up{adm} = \mathsf{filtering} \left( K\down{adm} \bm{\lambda}\up{read}\down{off} \right),
\end{equation}
where the filtering applies a deadband and output clamping to both linear and angular vector norms, thereby rejecting peak forces and inertial effects during motion. 
This effect minimizes interaction wrenches for disabled effectors, reducing collision forces during contact establishment and after contact removal, while also aiding in aligning feet with surface orientation. Implemented at input of the Retargeting level, this approach seamlessly integrates with operator command processing (\ref{eq:cmd_filtering}).

\section{Experimental Evaluation}

%This introductive paragraph is optional...
%Our experiments aim to demonstrate the following key capabilities: first, the controller effectively tracks desired wrench configurations through a combination of postural adjustments and joint flexibility modeling; second, the controller and effector admittance enable smooth and robust transitions between contact states; third, damping control law applied to wrenches efficiently mitigates oscillations induced by external perturbations; fourth, our robot can execute far-reaching, balance-challenging motions in multi-contact scenarios, demonstrating robustness against model inaccuracies; and finally, a more comprehensive analysis of robustness is conducted in simulation.

%We assessed our control pipeline on tasks involving wrench tracking, contact switching, damping oscillations arising from flexibility, and far-reaching with model errors. In simulation, we analyzed the robustness of our approach against motion speed and model errors. Additional multi-contact scenarios, including stair climbing and stepping on a sloped surface (Fig.~\ref{fig:concept}), are showcased in the attached video\footnote{Video \url{https://TODO}}.

%\begin{table}[t]
%\centering
%\caption{Observed Computing Times on Talos Robot in Milliseconds}
%\label{table:timing}
%\begin{tabular}{|ccccc|} 
%    \hline
%    Timing SEIKO & Median & $D_1$ ($$<10\%$$) & $D_9$ ($$>90\%$$) & Max\\
%    \hline
%    SEIKO Retargeting & $0.50$ & $0.49$ & $0.51$ & $0.56$\\
%    SEIKO Controller & $0.40$ & $0.39$ & $0.41$ & $0.43$\\
%    \hline
%\end{tabular}
%\end{table}

\subsection{Implementation Details}

We implemented SEIKO in C++ using RBDL (due to historical reasons) and Pinocchio \cite{carpentier2019pinocchio} rigid body libraries. More specifically, Pinocchio efficiently computes the analytical derivatives of the terms appearing in the equation (\ref{eq:diff_equilibrium}). We solve the QP problems using the QuadProg \cite{goldfarb1983numerically} solver.

The entire control pipeline operates at a frequency of $500$~Hz, with joint position commands interpolated at $2$~kHz before being transmitted to the robot's actuators. The median computing times observed on the internal computer of the Talos robot are $0.50$~ms and $0.40$~ms for SEIKO Retargeting and SEIKO Controller, respectively. The maximum measured times for each were $0.56$~ms and $0.43$~ms, respectively.

The Talos robot, manufactured by PAL Robotics, is a humanoid robot of $1.75$ m height with 32 DoFs. We measured with an independent weighing scale its actual total mass to be $99.7$ kg, while the URDF model provided by PAL assumes a mass of $93.4$ kg. This discrepancy of $6$ kg can be seen by the Force-Torque sensors in the feet, which enable our controller to adapt to this model error. We changed the robot's right hand and forearm with a 3D printed part that replaced the gripper and wrist joints beyond the elbow joint. The ball-shaped hand (point contact) allows us to apply high contact forces (up to $30$ kg) on the arm during multi-contact tests. After removing the right forearm joints and excluding the head joints, our QP solver works with $n = 25$ joints. All joints are used in position-controlled mode.

Throughout all our evaluations, we used as flexibility model $\bm{K}$ the position-control P gains imported from PAL's Gazebo simulation of the Talos robot. Unlike other work \cite{villa2022torque} that estimate precise flexibility model, our approach does not heavily depend on model accuracy. This is because our formulation with derivatives utilizes only the approximate ``gradient'' direction for whole-body control.

In all subsequent experiments, an expert operator issued velocity commands for each robot's effectors using dedicated 6-DoF input devices\footnote{3Dconnexion SpaceMouse: {\url{https://3dconnexion.com/uk/spacemouse/}}}, with one device assigned to each effector. Teleoperation was conducted with a clear, direct line of sight to the robot and its surrounding environment.

\begin{figure}[t]
    \centering
    \raisebox{0cm}{\includegraphics[trim=\trimscale{31}cm 0cm \trimscale{17}cm 0cm,clip,height=4.8cm]{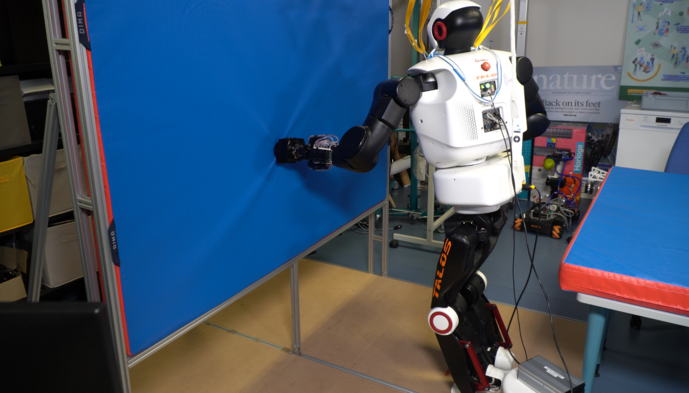}}
	\includegraphics[trim=0cm 0cm 0cm 0cm,clip,height=4.8cm]{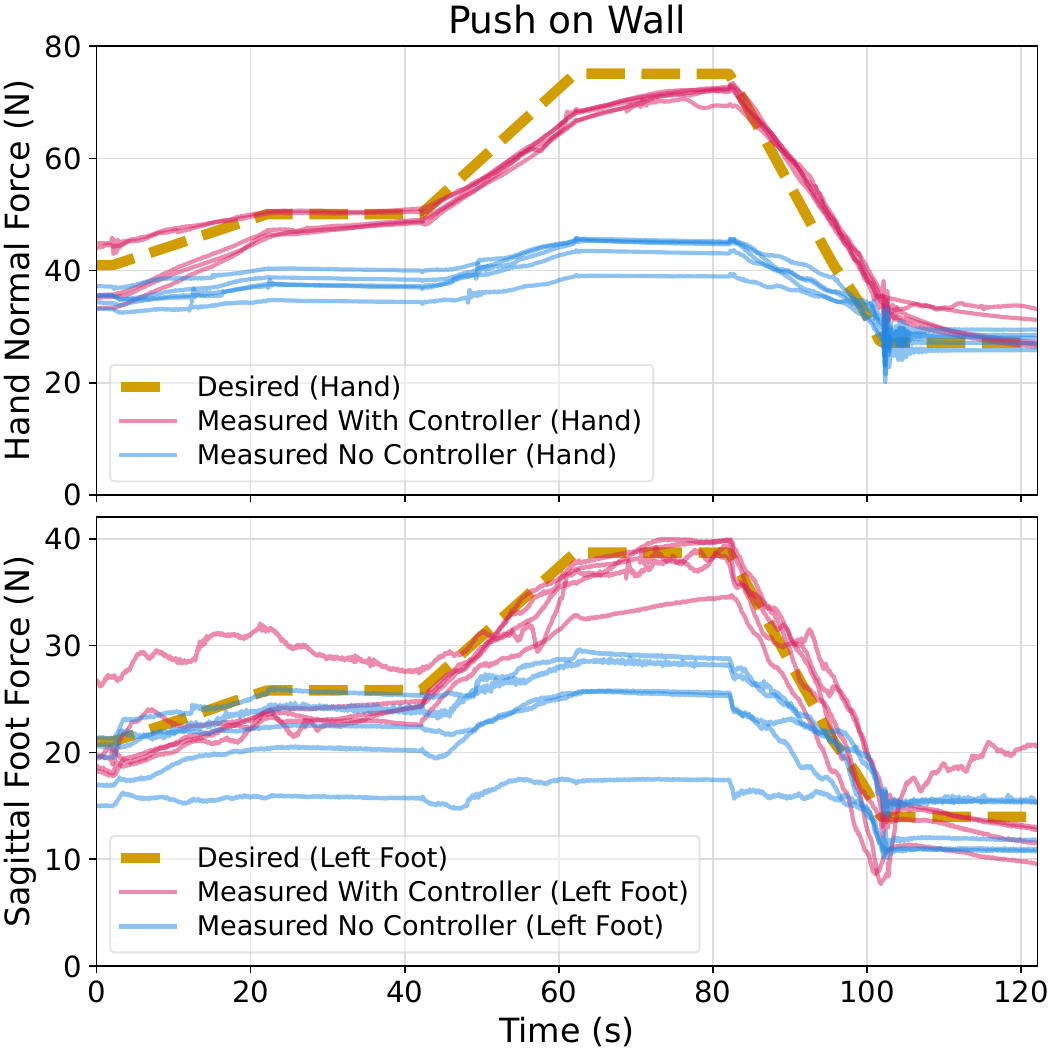}	
    \caption{Force distribution tracking during pushing tasks. The Talos robot (left) pushes a vertical wall using its left hand, following a predefined hand force target trajectory. Plots display the desired and measured normal force for the left hand (top) and the sagittal tangential force for the left foot (bottom); comparing with control enabled (5 trials) and without (5 trials).}
    \label{fig:push_wall}
\end{figure}

\subsection{Wrench Distribution Tracking}

In Fig.~\ref{fig:push_wall}, we illustrate the role of SEIKO Controller in realizing multi-contact wrench distribution during a hand pushing task. The robot initiates a point contact with a vertical wall using its left hand. The ``pushing mode'' of SEIKO is employed to command a target trajectory for the normal force applied on the wall. Retargeting adjusts the robot's posture slightly forward to apply a large force ($75$ N), and generates the desired contact wrenches, including opposing tangential forces on the feet in the sagittal plane.

We did not perform any identification or tuning of the robot flexibility model on the actual hardware, which may have significant errors. Estimating this flexibility \cite{villa2022torque} could enhance tracking accuracy, given that we observed near-perfect tracking performance in the Gazebo simulator which uses an ideal model.

The attached video\footnote{Additional videos: \url{https://hucebot.github.io/seiko_controller_website/}} demonstrates additional multi-contact scenarios, such as stair climbing and stepping on sloped surfaces (Fig.~\ref{fig:concept}). The observed motions of the robot are deliberately slow due to the focus on quasi-static movements. We also conducted additional comparisons with the prior method effector admittance control \cite{murooka2022centroidal} in \refsupp{Section~{\ref{sec:supp_comparison}}}.

\begin{figure}[t]
    \centering
    \raisebox{0cm}{\includegraphics[trim=\trimscale{36}cm 0cm \trimscale{25}cm 0cm,clip,height=4.6cm]{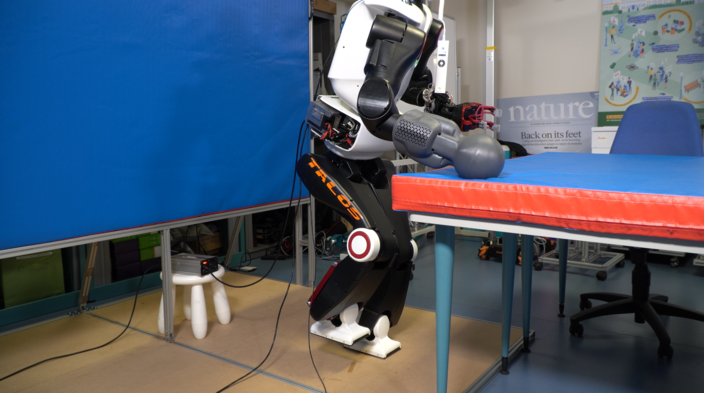}}
	\includegraphics[trim=0cm 0cm 0cm 0cm,clip,height=4.8cm]{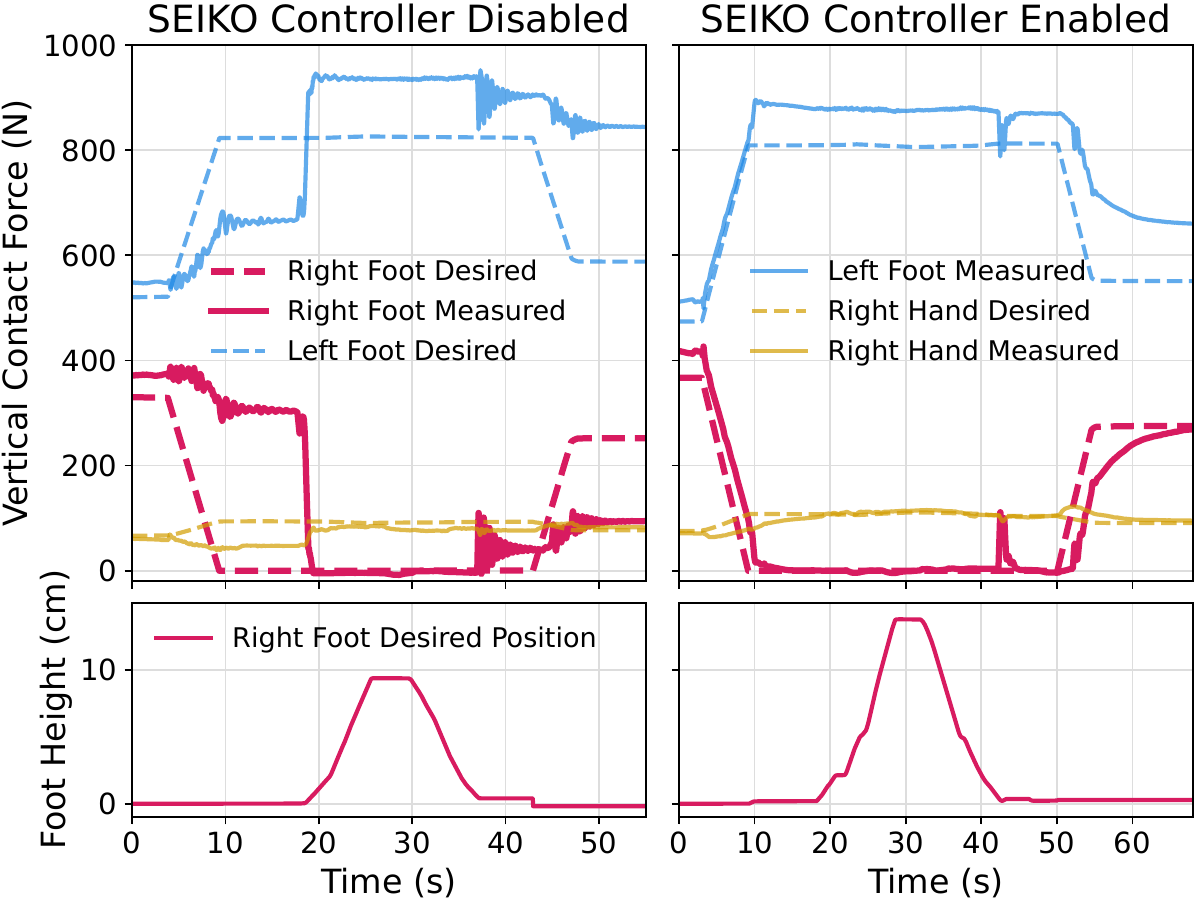}	
    \caption{Comparison of contact switch trials with and without SEIKO Controller. Initially, both feet and right hand are in contact. The operator teleoperated the robot to disable the right foot contact, lift the foot, and re-establish contact. Vertical contact forces $\bm{\lambda}\up{d}, \bm{\lambda}\up{read}$ (top row) and the desired vertical position of the right foot $\bm{X}\up{d}\down{right foot}$ (bottom row) are displayed.}
    \label{fig:foot_switch}
\end{figure}

\subsection{Contact Switch}

Fig.~\ref{fig:foot_switch} illustrates the foot contact switch capabilities, showcasing the Talos robot being teleoperated to lift and then re-establish contact with the right foot. Without the Controller, weight transfer from the right to the left foot and hand occurs abruptly during the foot lift. The robot did not fall as it was operating far from its feasibility boundaries. Conversely, when the controller and admittance scheme (equation (\ref{eq:admittance})) were enabled, the redistribution of contact wrenches became smooth and controlled. Additionally, at $t=43$ s, when the foot collided with the ground, the admittance control sightly lifted the foot to prevent unwanted ground forces before contact was re-established.

\begin{figure}[t]
    \centering
    \includegraphics[trim=4cm 0cm 0cm 0cm,clip,height=3.9cm]{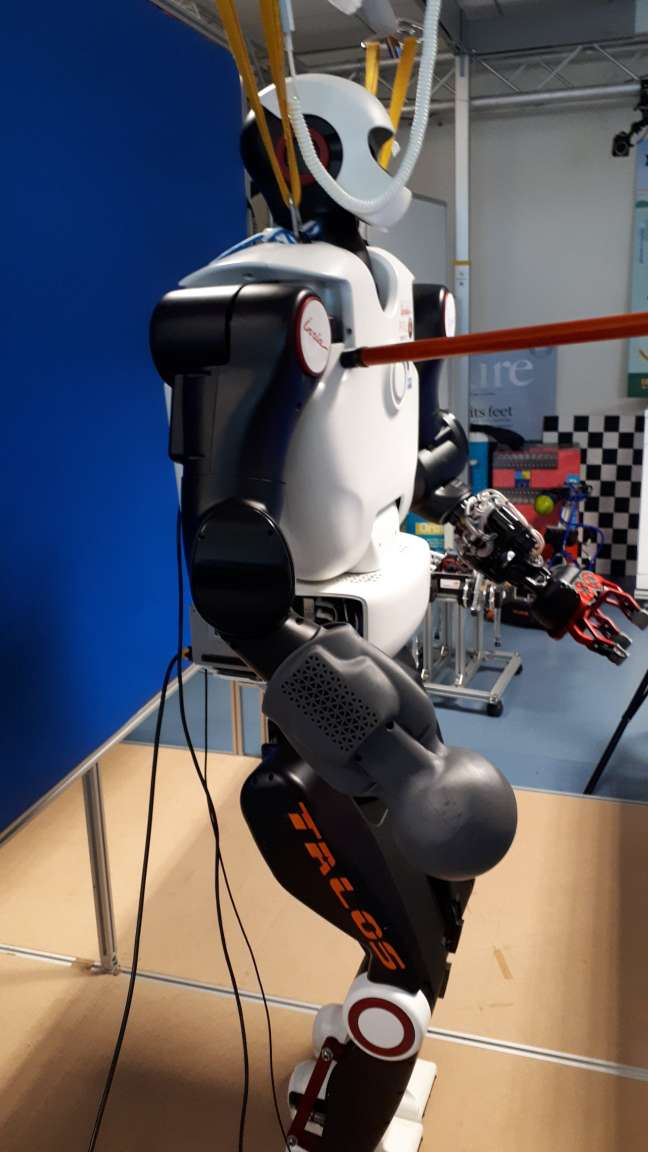}	
	\includegraphics[trim=0cm 0cm 0cm 0cm,clip,height=4.2cm]{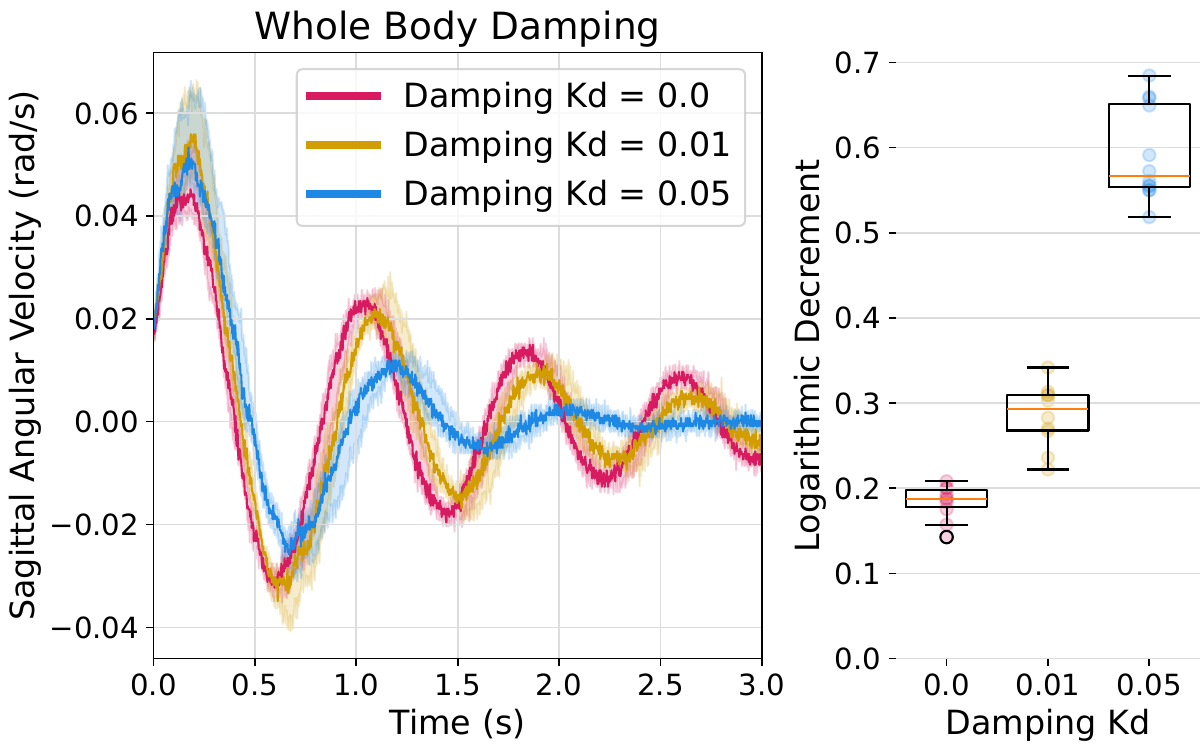}	
    \caption{Impact of damping gain $K_d$ on Talos's torso oscillations. Short pushes are applied (left), and IMU's gyroscope measures sagittal plane oscillation for varying $K_d$ values (middle). Damping effect is quantified using logarithmic decrement metric from oscillation peaks (right).}
    \label{fig:damping}
\end{figure}

\subsection{Whole Body Damping}

Imperfect stiff position control and flexibilities lead to small oscillations when disturbed, particularly noticeable on Talos in the sagittal plane, causing forward-backward oscillations. In equation (\ref{eq:feedback_law}), the controller's feedback law employs a damping term with the gain parameter $K_d$. %We show in Fig.~\ref{fig:damping} that \change{this feedback law on contact wrenches, which is the only feedback using measurement from sensors}, effectively attenuates these whole-body oscillations.
In Fig.~\ref{fig:damping}, we show that this feedback law applied to contact wrenches, serving as the only feedback mechanism in our formulation using sensor measurements, effectively attenuates the whole-body oscillations.

In double support, we applied short pushes (10-12 pushes, Fig.\ref{fig:damping} left) to the robot's torso and observed oscillations until energy dissipation. Using the controller, we tested various damping gain ($K_d = {0.0, 0.01, 0.05}$). We recorded unfiltered angular velocity in sagittal plane with pelvis IMU's gyroscope since it does not rely on model nor unobserved joint positions. Fig.~\ref{fig:damping} (center) shows median and $20\%-80\%$ deciles confidence interval of sagittal motion velocity. To quantify damping (Fig.~\ref{fig:damping} right), we estimated the averaged logarithmic decrement from oscillation peaks ($\delta = \mathsf{avg}\left(\log\left(\frac{\omega(t)}{\omega(t+T)}\right)\right)$), reflecting damping of oscillation amplitudes and linked to the damping ratio for under-damped systems.

In following experiments, the damping gain is set to $K_d = 0.02$, as higher values tended to be unstable near feasibility boundaries where model errors had a more pronounced effect.

\subsection{Far Reaching with Model Errors}

\begin{figure}[t]
    \centering
    \begin{minipage}{0.29\linewidth}
        \includegraphics[trim=\trimscale{17}cm 0cm \trimscale{30}cm 0cm,clip,width=0.95\linewidth]{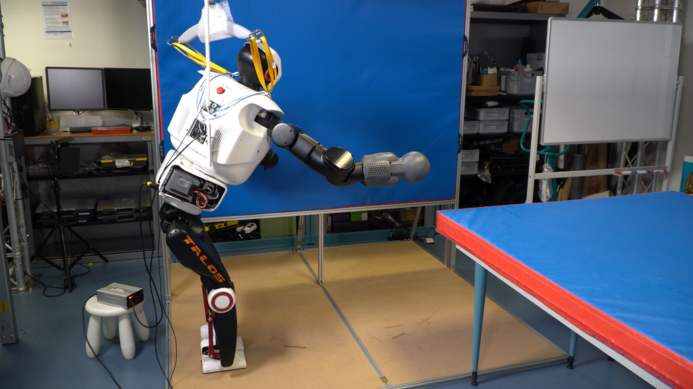}\\
        \includegraphics[trim=\trimscale{11}cm 0cm \trimscale{32}cm 0cm,clip,width=0.95\linewidth]{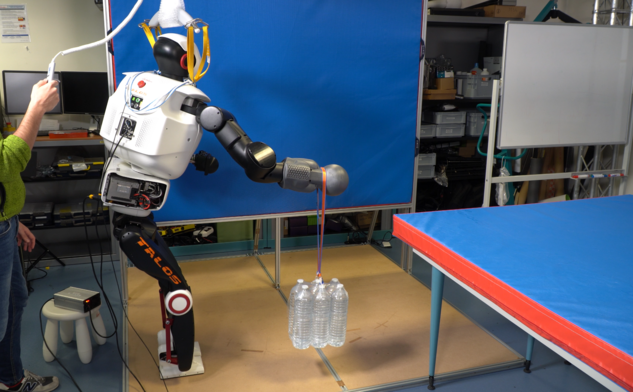}
    \end{minipage}%
    \begin{minipage}{0.69\linewidth}
    	\includegraphics[trim=0cm 0cm 0cm 0cm,clip,width=0.95\linewidth]{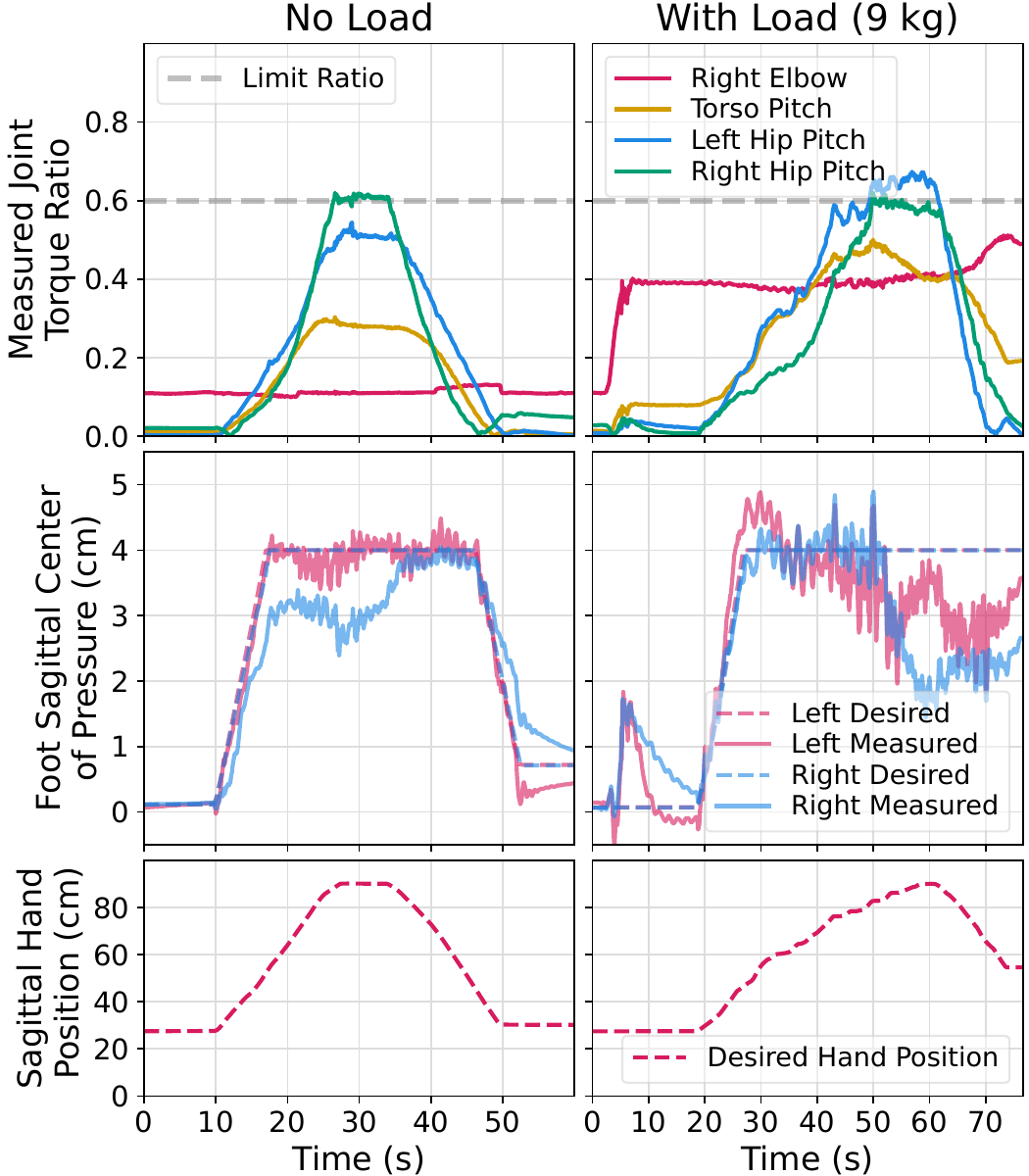}	
    \end{minipage}
    \caption{Far reaching task with and without adding a large unmodeled mass ($9$ kg) on the hand. The controller enforces joint torque ratio limits (top row, set to $0.6$) and tracks the foot contact wrenches (middle row) to ensure balance.}
    \label{fig:reaching_load}
\end{figure}

Fig.~\ref{fig:reaching_load} illustrates the capability of our approach to perform challenging far-reaching tasks near feasibility limits, even in the presence of large model errors. We teleoperated the right hand of the Talos robot for a forward-reaching motion as far as allowed by the controller, and added a $9$ kg load during operation on the hand to induce mass model errors. The robot remained stable thanks to the tracking of foot contact wrenches and adaptation of the whole-body posture. Additionally, the Controller through equation (\ref{eq:limit_joint_tau_adaptation}) prevents excessive violation of joint torques, with a limit ratio set to $\frac{|\tau\up{read}|}{\tau\up{max}} < 0.6$.

\begin{figure*}[t]
    \centering
    \includegraphics[height=4.1cm]{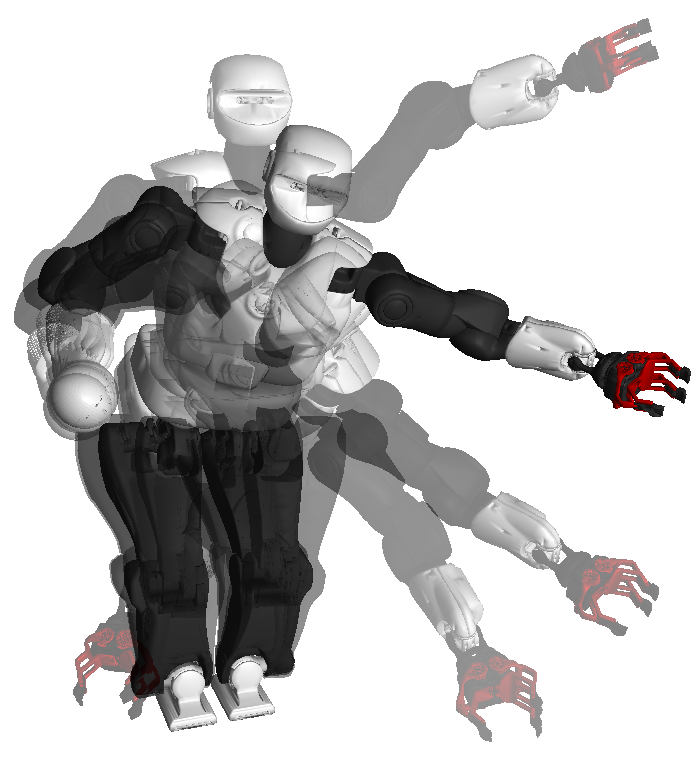}
	\includegraphics[trim=0cm 0cm 0cm 0cm,clip,height=4.2cm]{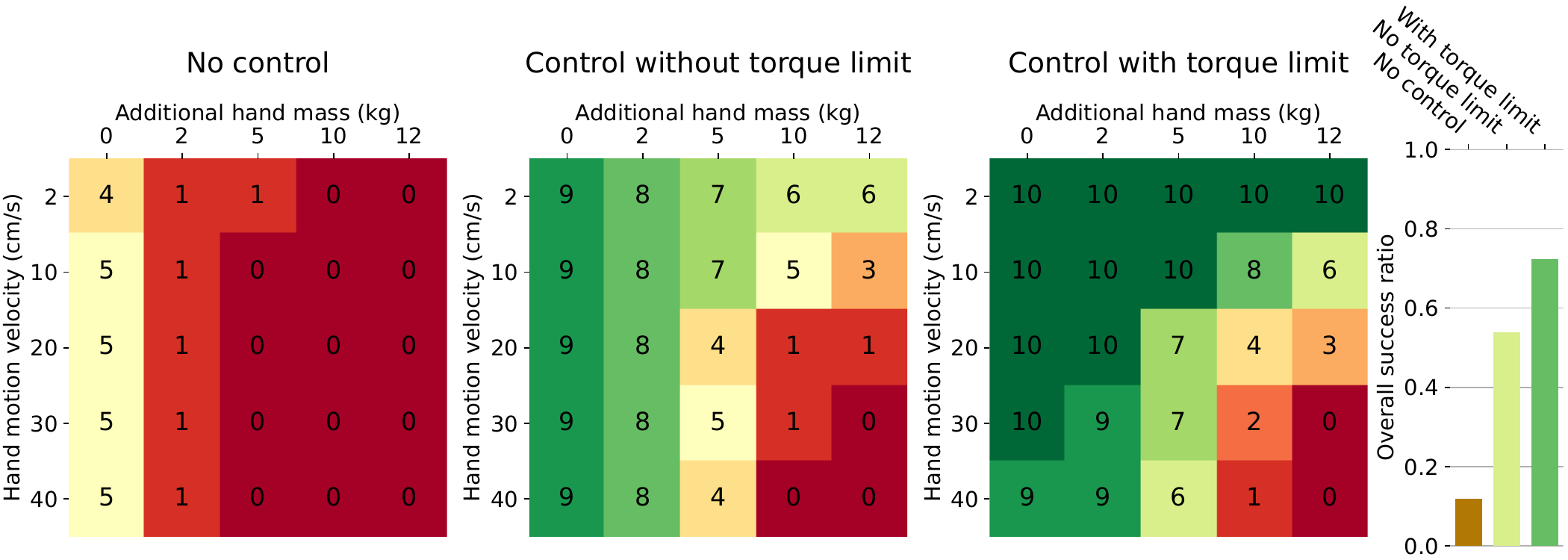}	
    \caption{Comparison of our controller's robustness against model errors and motion velocity. The Talos robot performs in double support $10$ far-reaching tasks at the edge of the feasibility boundary in the MuJoCo simulator (left). The number of successful trials without falling is indicated (out of $10$). Different combinations of hand motion velocity and added mass on the robot's hand are compared (middle). The comparison includes scenarios with the SEIKO controller disabled (only open-loop SEIKO retargeting), the SEIKO controller with only foot wrenches control, and the full controller also considering joint torque limits. Overall success ratio comparing the three controllers is given on right panel.}
    \label{fig:robustness_comparison}
\end{figure*}

\subsection{Robustness Evaluation}

We performed a comprehensive analysis of our approach's robustness using the MuJoCo simulator, as summarized in Fig.~\ref{fig:robustness_comparison}. The focus was on evaluating the impact of model errors and motion speed on system's balance. We simulated the Talos robot in double support, executing $10$ motion sequences reaching a distant target with the left hand and returning to the initial posture. The number of successful trials without fall for three conditions are reported: (i) without SEIKO Controller, (ii) with SEIKO Controller but without considering joint torque limits (\ref{eq:controller_ineq_tau}), (\ref{eq:limit_joint_tau_adaptation}), and (iii) using the full control method. Variations included hand Cartesian motion velocity (slow $2$ cm/s to fast $40$ cm/s) and additional mass on the left hand (none to $12$ kg).

We observed that MuJoCo's soft contact model produces a more pronounced flexibility behavior than Gazebo or even the actual robot. The presented results implicitly incorporate flexibility model errors, although they are not quantified.

SEIKO Retargeting without whole-body control (left) operates in open-loop and is partially robust to motion speed but struggles with model errors. Using SEIKO Controller (middle) significantly improves success rates, adapting joint position commands to handle additional hand mass for balance. However, unplanned posture adaptations and model errors near full extension reach actuator torque limits, leading to loss of control authority. Considering actuator torque limits in the controller (right) enhances robustness by optimizing posture and avoiding infeasible hand pose commands. Challenges persist at high speeds and heavy masses, where inertial effects violate the quasi-static assumption.

%\subsection{Full Teleoperation Scenario}
%-Make the impressive results first, the admittance and other after (basically in reverse order).
%-Explication slides: not focus on how it is done, but rather why/what, what is the problem and why. Add a slide of simplified assumption for expert (teleop, position controlled, quasi-static, optimization, very simple). Do no make a detailed presentation, but the video is an addition to the main paper.
%-On video, add annotation (since we have fixed point of view) or something to help the viewer to focus on what is interesting to watch (contact switch, weight transfer, admittance plane adaptation, etc...). May be not on all videos, only a few. Is in addition to potential subtitle.

\section{Discussion and Conclusion}

Our control architecture's robustness is showcased at moderate motion speeds (Fig.~\ref{fig:robustness_comparison}), but it inherently relies on the quasi-static assumption and is unsuitable for highly dynamic motions. Exploring more dynamic and agile motions is an avenue for future research. Establishing contact with stiff position-controlled robots requires precise and slow operator commands, even if effectors admittance (\ref{eq:admittance}) helps mitigating this problem. Future work could explore applying the proposed approach to robots using joint impedance control. As analyzed in \cite{villa2022torque}, we noted greater leg flexibility in the Talos robot than in our basic model. Although our controller enables successful contact transitions in teleoperated tasks, this significant difference hampers the quick contact switches needed for walking. Refining the flexibility model may allow walking capabilities.

The robot fell when attempting to climb large $20$ cm stairs for exceeding arm joint torque limits during the challenging contact switch. Despite being theoretically feasible according to the retargeting model, the adaptation of joint torque limits (\ref{eq:limit_joint_tau_adaptation}) is insufficient to ensure robustness if an infeasible contact transition is attempted due to model errors (e.g., underestimating the robot's weight).

%In our experiments, we commanded robot's effectors individually in velocity mode. Future work should aim to create an intuitive human-robot interface for complex multi-contact teleoperation. Additionally, for remote teleoperation, an interface providing sufficient feedback to the operator for accurate environmental awareness is essential. Our work does not address determining feasible contact sequences and stances, crucial for navigating complex terrain. Integrating assistance for the operator in this non-intuitive task should be part of the human-robot interface for practical teleoperation.

SEIKO Controller overcomes the inherent lack of direct control authority over contact forces of position-controlled by explicitly considering flexibilities. The whole-body multi-contact formulation is grounded in model and enhances robustness to moderate motion speeds and model errors, safely carrying substantial unmodelled loads at arm's length. The unified whole-body formulation employs a single feedback law on contact forces, effectively leveraging both postural change (i.e., CoM displacement) and contact force redistribution to regulate balance. Given that the primary advantage of humanoids and other multi-limbed robots lies in their strong versatility, this research paves the way for broadening the application and deployment of real-world scenarios, utilizing more capable and adaptable multi-contact systems in uncertain contexts and environments.

%%%
%%% Bibliography
%%%

\bibliographystyle{templates/IEEEtran}
\bibliography{references}

%%%
%%% Supplementary
%%%

\clearpage
\newpage
\makeatletter
\renewcommand \thesection{S\@arabic\c@section}
\renewcommand\thetable{S\@arabic\c@table}
\renewcommand \thefigure{S\@arabic\c@figure}
\setcounter{section}{0}
\setcounter{table}{0}
\setcounter{figure}{0}
\makeatother

\begin{figure*}[t]
\centering
\huge\textit{Supplementary Material:}\\
Multi-Contact Whole-Body Force Control\\for Position-Controlled Robots\\
\vspace{1em}
\large Quentin Rouxel, Serena Ivaldi, and Jean-Baptiste Mouret\\
\vspace{1em}
\end{figure*}

\appendix

\begin{figure*}[t]
    \centering
    \includegraphics[trim=0cm 0cm 0cm 0cm,clip,height=5.0cm]{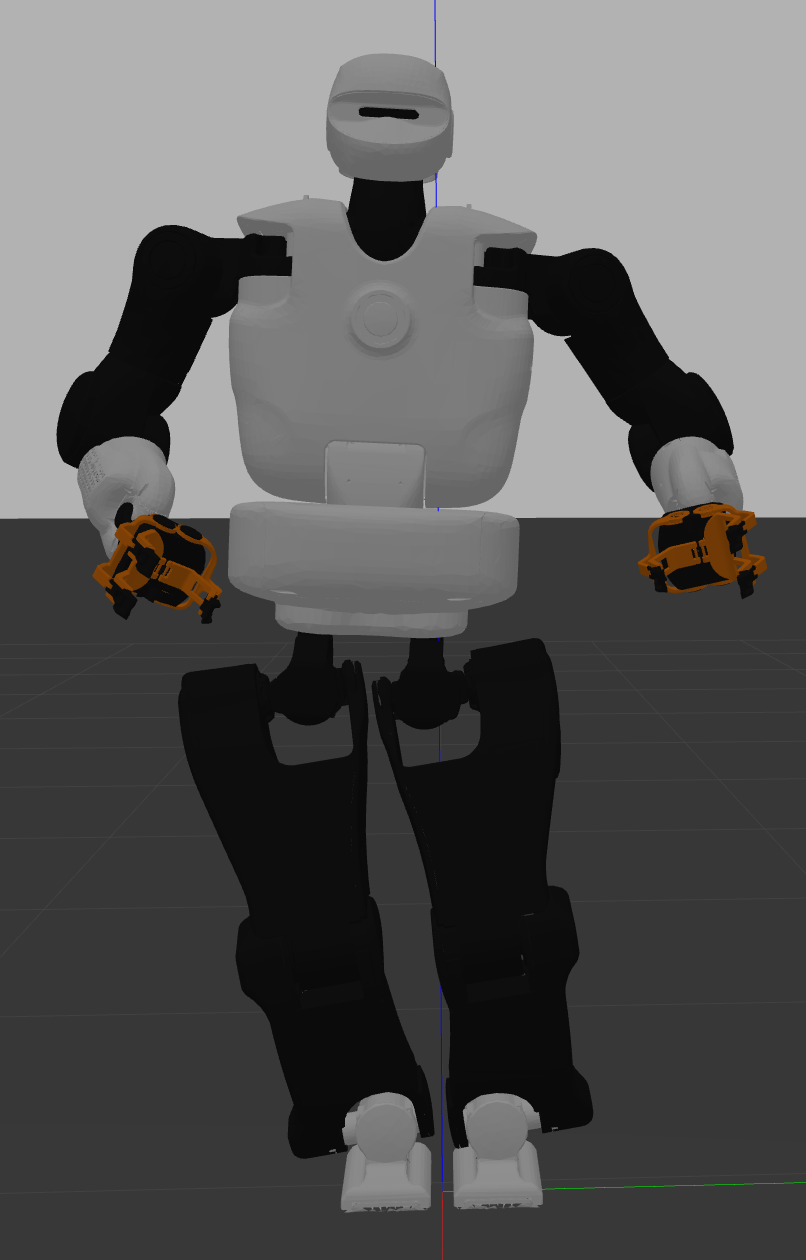}%
    \includegraphics[trim=0cm 0cm 0cm 0cm,clip,height=5.0cm]{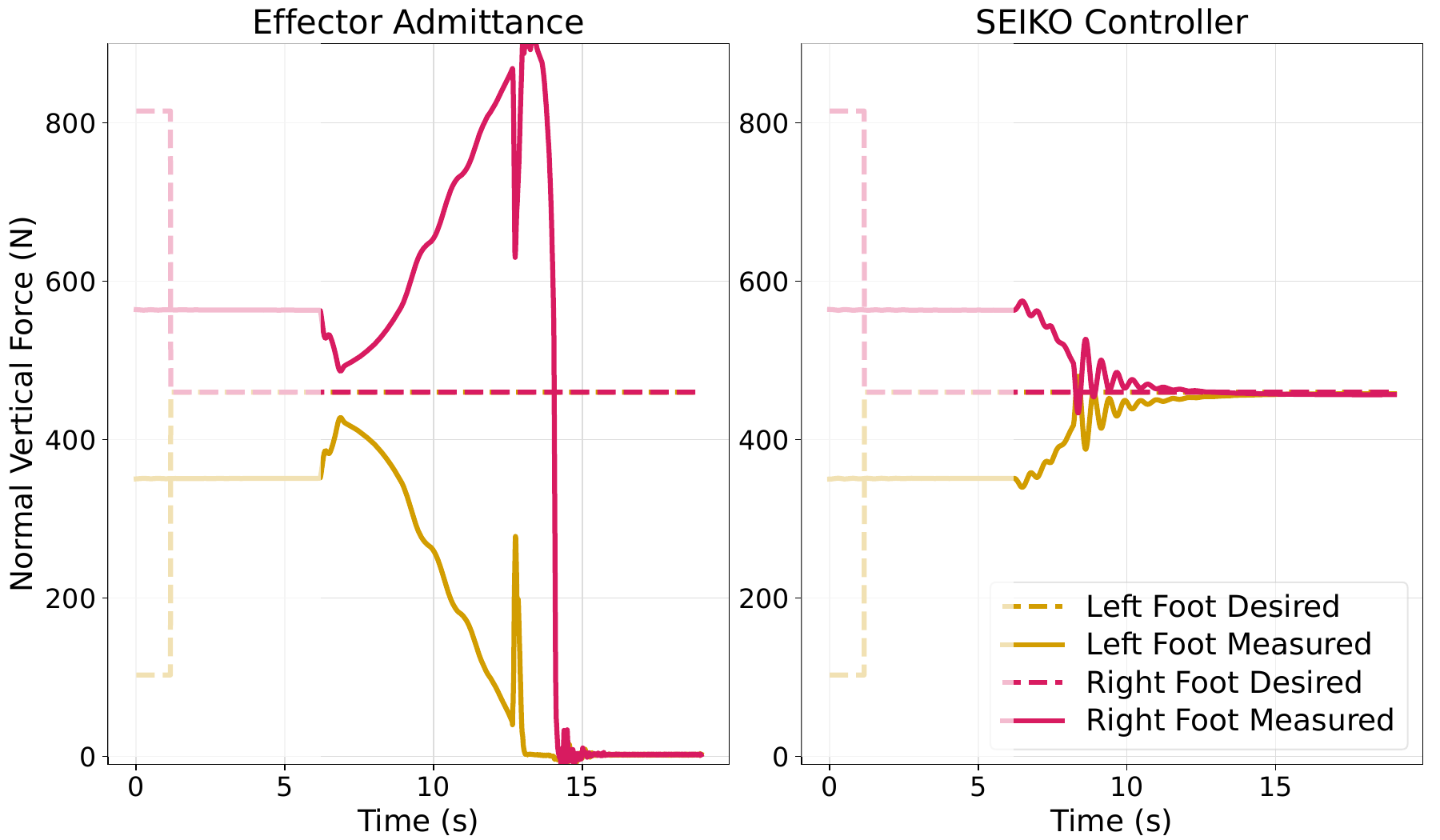}	
    \caption{Comparison of our SEIKO Controller and effector admittance \cite{murooka2022centroidal} for tracking inconsistent references on Talos humanoid robot simulated in double support using Gazebo simulator (left). SEIKO Retargeting is used to generate a configuration where most of the robot's weight is positioned above the right foot. At $t=1s$, the reference sent to the controller is overridden, requesting an equal weight distribution between the two feet, inconsistent with the desired posture and CoM position. Both controllers are initiated at $t=6s$ and foot normal force tracking is displayed for each controller (right). SEIKO Controller successfully tracks the overridden reference by shifting the CoM of the robot, while the effector admittance controller results in the robot falling.}
    \label{fig:baseline_foot}
\end{figure*}

\begin{figure*}[t]
    \centering
    \includegraphics[trim=0cm 0cm 0cm 0cm,clip,height=5.0cm]{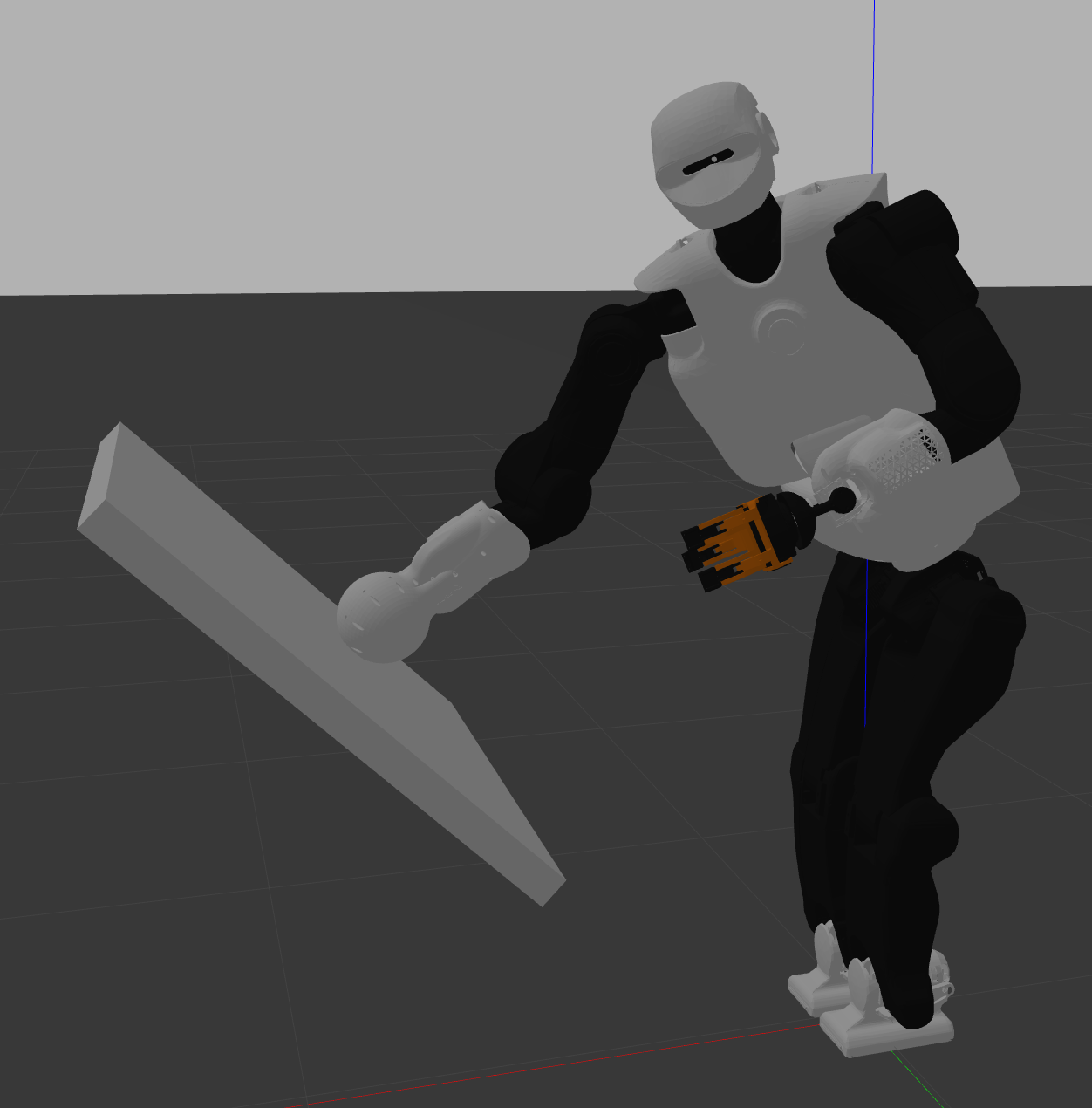}%
    \includegraphics[trim=0cm 0cm 0cm 0cm,clip,height=5.0cm]{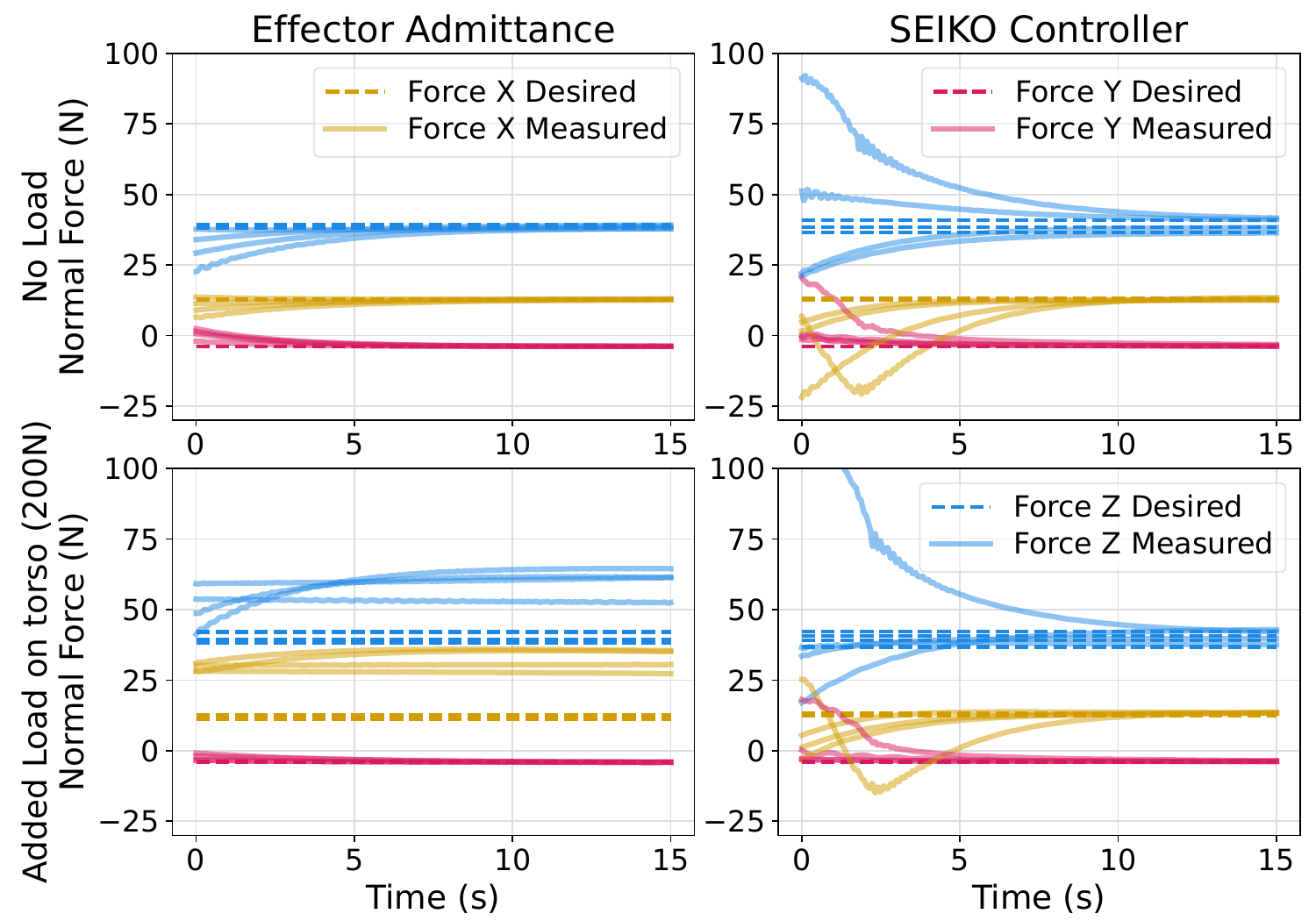}	
    \caption{Comparison between our SEIKO Controller and effector admittance \cite{murooka2022centroidal} for hand multi-contact and large model errors. Talos humanoid robot is simulated in Gazebo, with a posture featuring both feet and the right hand in contact (left). Tracking of the right hand force is compared across several initial contact forces (right). In the second row, a large external vertical force ($200N$) is applied on the robot's torso. The effector admittance scheme fails to track the reference when faced with large external disturbances.}
    \label{fig:baseline_hand}
\end{figure*}

%\begin{myhideenv}

\subsection{Comparison With Effector Admittance Control}\label{sec:supp_comparison}

As detailed in Section~\ref{sec:related_works}, prior studies focusing on position-controlled robots \cite{caron2019stair, cisneros2019qp, samadi2021humanoid} often use the method introduced in \cite{kajita2010biped} for regulating contact forces. These approaches named ``effector admittance'', ``foot force difference control'', or ``damping control for limb ends'' all operate on a similar principle. 
They employ an admittance feedback law applied to each effector of the robot, which adjusts its Cartesian pose reference based on desired and measured contact forces before being realized via Inverse Kinematics (IK).

To compare our method against this baseline, we specifically implemented the ``damping control'' approach described in \cite{murooka2022centroidal}, Section~IV.C. We substituted the SEIKO Controller block in Fig.~\ref{fig:architecture} with an IK module. Prior to the IK calculation, we apply the following admittance law to each enabled and disabled effector:
\begin{align}
    & \bm{X}\up{IK}(t) = \bm{X}\up{d}(t) \oplus \Delta\bm{r}(t),\\
    & \Delta\bm{r}(t+\Delta t) = \Delta\bm{r}(t) \oplus
    \Delta t \left(K\down{f}(\tilde{\bm{\lambda}}\up{read} - \bm{\lambda}\up{d}) \ominus K\down{s}\Delta\bm{r}(t)\right),
\end{align}
where $\bm{X}\up{IK}(t) \in \SE$ is the corrected effector pose input sent to the IK, $\Delta\bm{r}(t) \in \SE$ is the computed admittance pose offset (slightly abusing the $\SE$ notation), and $K\down{f}, K\down{s} \in \R$ are the manually tuned wrench and spring gains, respectively.
It is worth noting that we did not implement for our comparison the two additional feedback effects described in \cite{murooka2022centroidal}, as detailed in Sections~IV.A and B. These two effects only use position measurements to regulate the centroidal pose. However, our Talos robot is controlled in position with stiff gains, and no significant position errors can be measured with respect to the desired posture optimized by our SEIKO Retargeting (posture flexibility is unobservable on joint sensors).

From a theoretical standpoint, contact forces in rigid multi-contact scenarios are influenced by two key factors. Firstly, the distribution of forces among different contacts can be selected within the redundant contact nullspace without inducing motion. Secondly, this nullspace is defined by the posture, specifically the position of the CoM. As its formulation is not whole-body, the effector admittance approach only utilizes the first component of force redistribution, while our SEIKO Controller offers a unified formulation capable of leveraging both force redistribution and postural adjustments.

For instance, the effector admittance approach relies on the assumption that retracting the effector away from the surface will decrease the contact force. While this assumption holds true in many scenarios, it can lead to failure when confronted with significant model errors, flexibility, or challenging postures that bring the robot's configuration close to its feasibility limits. The following simulation experiments highlight two cases where the effector admittance method fails to accurately track the force reference due to its inability to account for whole-body postural adjustments.

\subsubsection{Inconsistent Reference Input}

Fig.~\ref{fig:baseline_foot} presents a clear example of a simulated double support scenario where retracting the foot position away from the surface fails to decrease the contact force. This situation arises when the planner module (in this case, SEIKO Retargeting) outputs an erroneous desired reference, subsequently sent to the underlying controller. Due to the inconsistency between the desired contact force and the desired posture, the controller is unable to track both simultaneously. The effector admittance control attempts to lift the right foot, which bears most of the robot's weight, leading to the robot's fall. In contrast, our SEIKO Controller prioritizes balance by focusing on tracking the contact forces and adjusting the posture through the whole-body formulation.

\subsubsection{Large Model Errors}

In numerous instances where the robot's configuration remained away from the feasibility limits, we observed that the effector admittance method effectively regulated the contact forces. However, the robustness of the controller is challenged in scenarios with large model errors. Fig.~\ref{fig:baseline_hand} illustrates a failure case where a substantial external force applied to the torso necessitates the adaptation of the robot's posture to accurately track the desired contact forces, a capability only exhibited by the SEIKO Controller.

\subsection{Derivative of Equilibrium Equations}\label{sec:supp_equations}

In the quasi-static case, the whole-body equilibrium equation in joint space is expressed as:
\begin{equation}
    \bm{g}(\bm{q}) = \bm{S}\bm{\tau} + \bm{J}(\bm{q})^\T \bm{\lambda}.
\end{equation}
Since this equation is nonlinear with respect to $\bm{q}$, we linearize it by approximating its derivative through consideration of small variations in the configuration
$\left(\bm{q}+\Delta\bm{q}, \bm{\lambda}+\Delta\bm{\lambda}, \bm{\tau}+\Delta\bm{\tau}\right)$:
\begin{equation}
    \bm{g}(\bm{q}+\Delta\bm{q}) = \bm{S}(\bm{\tau}+\Delta\bm{\tau}) + \bm{J}(\bm{q}+\Delta\bm{q})^\T (\bm{\lambda} + \Delta\bm{\lambda}),
\end{equation}
which is linearized and approximated as following considering partial derivatives of the gravity vector and the contact Jacobian matrix:
\begin{equation}
\begin{aligned}
    \bm{g}(\bm{q}) + \fracdiff{\bm{g}}{\bm{q}}(\bm{q})\Delta\bm{q} =
    & \bm{S}\bm{\tau} + \bm{S}\Delta\bm{\tau}
    + \bm{J}(\bm{q})^\T\bm{\lambda} + \bm{J}(\bm{q})^\T\Delta\bm{\lambda} \\
    & + \left(\fracdiff{\bm{J}}{\bm{q}}^\T(\bm{q})\bm{\lambda}\right)\Delta\bm{q}
    + \left(\fracdiff{\bm{J}}{\bm{q}}^\T(\bm{q})\Delta\bm{\lambda}\right)\Delta\bm{q}.
\end{aligned}
\end{equation}
We further neglect the following term, considering it to be of second order: 
\begin{equation}
    \left(\fracdiff{\bm{J}}{\bm{q}}^\T\Delta\bm{\lambda}\right)\Delta\bm{q} \approx \bm{0},
\end{equation}
which leads to the derivative-based linear approximation of the equilibrium equation:
\begin{equation}\label{eq:supp_diff_equilibrium}
\begin{aligned}
    \bm{g}(\bm{q}) + \fracdiff{\bm{g}}{\bm{q}}(\bm{q})\Delta\bm{q} =
    & \bm{S}\bm{\tau} + \bm{S}\Delta\bm{\tau}
    + \bm{J}(\bm{q})^\T\bm{\lambda} + \bm{J}(\bm{q})^\T\Delta\bm{\lambda} \\
    & + \left(\fracdiff{\bm{J}}{\bm{q}}^\T(\bm{q})\bm{\lambda}\right)\Delta\bm{q}.
\end{aligned}
\end{equation}

Note that if we choose instead to linearize the equilibrium equation by only considering small variations in the posture $\left(\bm{q}+\Delta\bm{q}, \bm{\lambda}, \bm{\tau}\right)$, then the following term appears:
\begin{equation}\label{eq:supp_diff_bilinear_term}
    \left(\frac{\partial \bm{J}}{\partial \bm{q}}^{T}(\bm{q})\bm{\lambda}\right)\Delta \bm{q},
\end{equation}
which is of first order but also bilinear with respect to the decision variables $\left(\Delta\bm{q}, \bm{\lambda}\right)$. However, bilinear terms cannot be expressed in a QP formulation. Therefore, instead of neglecting the term (\ref{eq:supp_diff_bilinear_term}), which is of first order, the former expression (\ref{eq:supp_diff_equilibrium}) provides a better approximation of the equilibrium equation by considering $\left(\Delta\bm{q}, \Delta\bm{\lambda},\Delta\bm{\tau}\right)$ as the decision variables.

In the formulation of SEIKO Retargeting, the equality constraint (\ref{eq:retargeting_eq_equilibrium}) is obtained by setting $\bm{q} = \bm{q}\up{d}$ and $\bm{\lambda} = \bm{\lambda}\up{d}$ in (\ref{eq:supp_diff_equilibrium}).

In the formulation of SEIKO Controller, the derivative of the joint flexibility model (\ref{eq:diff_flexiblity}) is combined with (\ref{eq:supp_diff_equilibrium}) by setting $\bm{q} = \bm{q}\up{flex}, \bm{\lambda} = \bm{\lambda}\up{flex}, \bm{\tau} = \bm{\tau}\up{flex}$. This results in the derivative-based linear approximation of the equilibrium under flexibility:
\begin{equation}
\begin{aligned}
    & \bm{g}(\bm{q}\up{flex}) + \fracdiff{\bm{g}}{\bm{q}}(\bm{q}\up{flex})\Delta\bm{q}\up{flex} = 
    ~\bm{S}\bm{\tau}\up{flex}\\
    &+ \bm{S}\bm{K}\Delta\bm{\theta}^{\text{cmd}} - \bm{S}\bm{K}\bm{S'}\Delta\bm{q}\up{flex}\\
    &+ \bm{J}(\bm{q}\up{flex})^\T\bm{\lambda}\up{flex} + \bm{J}(\bm{q}\up{flex})^\T\Delta\bm{\lambda}\up{flex}\\
    &+ \left( \fracdiff{\bm{J}}{\bm{q}}^\T(\bm{q}\up{flex})\bm{\lambda}\up{flex} \right) \Delta\bm{q}\up{flex},
\end{aligned}
\end{equation}
yielding equation (\ref{eq:diff_equilibrium_flexiblity}).

\subsection{Whole-Body Balance Conditions}\label{sec:supp_balance_conditions}

Ensuring the robustness and stability of the overall system requires sending feasible desired configurations to the controller, taking into account the robot's balance to prevent potential falls. Balance sufficient conditions such as the projection of the CoM onto the support polygon (for flat ground) or the Gravito-Inertial Wrench Cone (GIWC) \citesupp{caron2015leveraging} are advantageous for reduced models like the centroidal model, as they enable faster computations. 

But these conditions do not consider the whole-body state and thus overlook joint kinematic and actuator torque limits. Instead, our formulation considers the complete whole-body configuration in quasi-static equilibrium. In this case, considering the CoM is not necessary, and both our SEIKO Retargeting and Controller formulation deliberately never explicitly include CoM consideration. 

A sufficient condition for the robot to be balanced is that all contacts remain asymptotically stable, i.e., in contact with the environment without any relative motion between the effector in contact and the surface. For these conditions to be satisfied, both in static and dynamic cases, it is sufficient that for each contact, the contact wrench remains within a 6d or 3d convex polytope (see Section~\ref{sec:supp_contact_conditions}). These conditions are explicitly enforced in SEIKO Retargeting QP by the inequality constraints in equation (\ref{eq:retargeting_ineq_contact}).

Given our quasi-static assumption and instantaneous formulation (without considering the future), enforcing conditions at the next time step is sufficient to meet the asymptotic requirement. If these conditions hold true at the next time step, they should theoretically extend into the future. Furthermore, the quasi-static equilibrium assumption is enforced by ensuring that the computed posture and contact forces satisfy the equilibrium equation (\ref{eq:equilibrium}) through its linear approximation in equations (\ref{eq:diff_equilibrium}) and (\ref{eq:diff_equilibrium_flexiblity}).

\subsection{Contact Stability Conditions}\label{sec:supp_contact_conditions}

A plane contact, which constrains 6 DoFs, between the robot's effector and the environment is considered stable when the relative linear and angular velocities between the effector and the contact surface are zero. Similarly, for point contact (3 DoFs), only the linear velocity is taken into account. For stability to be ensured, a sufficient condition is that the forces and torques, denoted by $\bm{\lambda} = \mat{f_x & f_y & f_z & \tau_x & \tau_y & \tau_z}^\T$, exerted on the environment by this contact, must satisfy the following convex inequalities \citesupp{caron2015stability} (the unilaterally, friction pyramid and Center of Pressure (CoP) constraints):
\begin{align}
& f_z \geq 0, \\
& \left|\frac{f_x}{f_z}\right| \leq \mu,~~ \left|\frac{f_y}{f_z}\right| \leq \mu, \\
& \left|\frac{\tau_y}{f_z}\right| \leq CoP\down{x}\up{max},~~ \left|\frac{\tau_x}{f_z}\right| \leq CoP\down{y}\up{max}, \\
& \tau_z\up{min} \leq \tau_z \leq \tau_z\up{max},
\end{align}
where $\mu, CoP\down{x}\up{max}, CoP\down{y}\up{max} \in R$ are respectively the friction coefficient, and the maximum half lengths in the $X$ and $Y$ directions of the rectangular contact surface. The local $Z$ axis is aligned with the contact normal. Please refer to \citesupp{caron2015stability} for the expression of $\tau_z\up{min}$ and $\tau_z\up{max}$. 

Additionally, a bound on the maximum normal force $f\up{max} \in \R$ can be included to prevent excessive force from being applied, particularly on a fragile effector:
\begin{equation}
    f_z \leq f\up{max}.
\end{equation}

These inequality equations can be linearly rewritten as (refer to \citesupp{feng2015optimization} for details):
\begin{equation}
    \bm{C}\down{contact}\bm{\lambda} + \bm{c}\down{contact} \geq \bm{0},
\end{equation}
where $\bm{C}\down{contact} \in \R^{18 \times 6}$ and $\bm{c}\down{contact} \in \R^{18}$ for a plane contact, and $\bm{C}\down{contact} \in \R^{6 \times 3}$ and $\bm{c}\down{contact} \in \R^{6}$ for a point contact.

In (\ref{eq:retargeting_ineq_contact}), all the contacts are stacked and $\bm{\lambda}$ is replaced with $\bm{\lambda}\up{d}+\Delta\bm{\lambda}\up{d}$.

\subsection{Improved Contact Switching Procedure}\label{sec:supp_contact_switch}

We enhanced the contact switching procedure compared to our prior SEIKO Retargeting work \cite{seiko}. To remove a contact, the desired contact force must smoothly be reduced to zero, often necessitating a smooth adjustment of the whole-body posture. To address this, SEIKO Retargeting employs a strategy of increasing the weight associated with the regularization task (\ref{eq:retargeting_cost_lambda}): by heavily penalizing the contact force, the whole-body optimization naturally diminishes the force and adjusts the posture accordingly.

In \cite{seiko}, we empirically observed that smoothly increasing the weight of task (\ref{eq:retargeting_cost_lambda}) using exponential growth yielded satisfactory switching motions. The contact is removed from the formulation when the desired force falls below a small threshold. However, the exponential choice and tuning of this procedure were arbitrary and not based on the robot's physical limits.

In this letter, we propose a refinement of this procedure. Instead of gradually increasing the weight of task (\ref{eq:retargeting_cost_lambda}), we set it instantaneously to a high value (typically $1e4$). The smoothness of the transition is ensured by introducing two inequality constraints on the maximum rates of change for joint positions $\dot{\bm{\theta}}\up{max}$ (\ref{eq:retargeting_ineq_vel_joint}) and contact wrenches $\dot{\bm{\lambda}}\up{max}$ (\ref{eq:retargeting_ineq_vel_lambda}). 

While the overall generated motions between the two procedures are very similar, this updated approach requires fewer arbitrary choices and tuning. Additionally, it offers the added benefit of explicit control over transition duration by setting the rate of change limits, which has clear physical semantics.

\bibliographystylesupp{templates/IEEEtran}
\bibliographysupp{references}

%\end{myhideenv}
\end{document}